\documentclass{article}

\usepackage{caption}
\usepackage{microtype}
\usepackage{graphicx}
\usepackage{subcaption}
\usepackage{booktabs} % for professional tables

\usepackage{hyperref}
\usepackage[breakable]{tcolorbox}

\usepackage{amssymb, amsmath}
\usepackage{siunitx}
\usepackage{bm}
\usepackage{pifont}
\definecolor{Gray}{gray}{0.9}
\definecolor{gr}{RGB}{0, 146, 0}

\definecolor{orange}{HTML}{ff9900} %orange

\definecolor{green}{HTML}{34a853}
\definecolor{lightgreen}{HTML}{b6d7a8}
\definecolor{seagreen}{HTML}{3CB371}

\definecolor{lightgray1}{HTML}{d9d9d9}
\definecolor{lightyellow2}{HTML}{ffe599}

\definecolor{lightblue}{HTML}{9fc5e8}
\definecolor{lightcornflowerblue3}{HTML}{c9daf8}

\definecolor{purple}{HTML}{9900ff} 
\definecolor{lightpurple1}{HTML}{8e7cc3}
\definecolor{lightpurple}{HTML}{b4a7d6}
\definecolor{lightred}{HTML}{e06666}

\usepackage{soul}
\colorlet{exqcolor}{lightgreen!70}
\colorlet{exccolor}{lightgray1!55}
\colorlet{exdcolor}{orange!30}
\colorlet{exacolor}{lightblue!90}
\colorlet{execolor}{lightred!50}
\colorlet{exncolor}{lightpurple!50}

\usepackage{multirow}
\usepackage{booktabs}
\usepackage{graphicx}
\usepackage{tabularx}
\usepackage{adjustbox}
\usepackage{array}
\usepackage{pifont}
\usepackage{xspace}
\usepackage{tcolorbox}
\usepackage{colortbl}
\usepackage{hyperref}
\usepackage{arydshln}
\usepackage{wrapfig}
\usepackage{soul}
\usepackage{makecell}
\usepackage{color, xcolor}
\usepackage{pgfplots}
\definecolor{tiffanyblue}{RGB}{129,216,208}
\definecolor{bangdiblue}{RGB}{0,149,182}
\definecolor{kleinblue}{RGB}{0,47,167}
\usepackage{amssymb}
\usepackage{circledtext}
\usetikzlibrary{shapes}
\usetikzlibrary{arrows,decorations.pathmorphing,backgrounds,positioning,fit,petri}
\usetikzlibrary{arrows.meta,fit,shapes.arrows}
\usetikzlibrary{positioning,shadows,patterns}
\usepackage{tikz}
\usepackage{pgfplots}
\usepackage{xspace}

\newcommand{\ours}{\textsc{Golf}\xspace}
\newcommand{\eg}{\emph{e.g.}}

\usepackage[ruled,vlined,linesnumbered]{algorithm2e}
\SetKwInput{KwInput}{Input}
\SetKwInput{KwOutput}{Output}
\SetKw{KwTo}{to}

\newtcolorbox{promptbox}[2][]{
	width=\linewidth,
	colback = gray!8, 
	colframe = gray!8, 
	boxsep=0pt,left=5pt,right=5pt,top=2pt,bottom=2pt,
	fontupper=\linespread{1.2}\selectfont,
	title=#2,#1,
        fontupper=\small}
        
\makeatletter
\def\adl@drawiv#1#2#3{%
        \hskip.5\tabcolsep
        \xleaders#3{#2.5\@tempdimb #1{1}#2.5\@tempdimb}%
                #2\z@ plus1fil minus1fil\relax
        \hskip.5\tabcolsep}
        
\newcommand{\cdashlinelr}[1]{%
  \noalign{\vskip\aboverulesep
           \global\let\@dashdrawstore\adl@draw
           \global\let\adl@draw\adl@drawiv}
  \cdashline{#1}
  \noalign{\global\let\adl@draw\@dashdrawstore
           \vskip\belowrulesep}}
\makeatother
\usepackage[preprint]{icml2026}

\usepackage{fvextra}

\DefineVerbatimEnvironment{verbatim}{Verbatim}{
  breaklines=true,
  breakanywhere=true,
  breaksymbolleft={},
}

\tcbset{
  promptbox/.style={
    colback=gray!5!white,        % very light gray background
    colframe=black!50,           % soft black frame
    coltitle=black,              % black title text
    colbacktitle=gray!10,        % slightly darker title background
    fonttitle=\bfseries,
    boxrule=0.4pt,
    arc=1mm,
    left=1mm,
    right=1mm,
    top=1mm,
    bottom=1mm,
    title={},
  }
}

\usepackage{amsmath}
\usepackage{amssymb}
\usepackage{mathtools}
\usepackage{amsthm}
\usepgfplotslibrary{fillbetween}

\usepackage[capitalize,noabbrev]{cleveref}

\theoremstyle{plain}

\theoremstyle{definition}

\theoremstyle{remark}

\usepackage[disable,textsize=tiny]{todonotes}
\icmltitlerunning{Bootstrapping Exploration with Group-Level Natural Language Feedback in Reinforcement Learning}

\begin{document}

\twocolumn[
  \icmltitle{Bootstrapping Exploration with Group-Level \\ Natural Language Feedback in Reinforcement Learning}



  \icmlsetsymbol{intern}{*}
  \icmlsetsymbol{corresponding}{†}
  
  \begin{icmlauthorlist}
    \icmlauthor{Lei Huang}{sch,comp,intern}
    \icmlauthor{Xiang Cheng}{comp}
    \icmlauthor{Chenxiao Zhao}{comp}
    \icmlauthor{Guobin Shen}{comp}
    \icmlauthor{Junjie Yang}{comp}
    \icmlauthor{Xiaocheng Feng}{sch,corresponding}
    \icmlauthor{Yuxuan Gu}{sch}
    \icmlauthor{Xing Yu}{comp,corresponding}
    \icmlauthor{Bing Qin}{sch}
  \end{icmlauthorlist}

  \icmlaffiliation{sch}{Harbin Institute of Technology}
  \icmlaffiliation{comp}{Xiaohongshu Inc}

  \icmlcorrespondingauthor{Xing Yu}{yuanshan2@xiaohongshu.com}
  \icmlcorrespondingauthor{Xiaocheng Feng}{xcfeng@ir.hit.edu.cn}


  \vskip 0.3in
]



\printAffiliationsAndNotice{$^*$Work done during internship at Xiaohongshu. $^\dagger$Corresponding authors.}  % no special notice (required even if empty)

\begin{abstract}
Large language models (LLMs) typically receive diverse natural language (NL) feedback through interaction with the environment.
However, current reinforcement learning (RL) algorithms rely solely on scalar rewards, leaving the rich information in NL feedback underutilized and leading to inefficient exploration.
In this work, we propose \ours, an RL framework that explicitly exploits \textbf{G}r\textbf{O}up-level \textbf{L}anguage \textbf{F}eedback to guide targeted exploration through actionable refinements. 
\ours aggregates two complementary feedback sources: (i) \emph{external critiques} that pinpoint errors or propose targeted fixes, and (ii) \emph{intra-group attempts} that supply alternative partial ideas and diverse failure patterns. 
These group-level feedbacks are aggregated to produce high-quality refinements,   which are \emph{adaptively} injected into training as off-policy scaffolds to provide targeted guidance in sparse-reward regions.
Meanwhile, \ours jointly optimizes generation and refinement within a unified RL loop, creating a virtuous cycle that continuously improves both capabilities.
Experiments on both verifiable and non-verifiable benchmarks show that \ours achieves superior performance and exploration efficiency, achieving 2.2$\times$ improvements in sample efficiency compared to RL methods trained solely on scalar rewards.
Code is available at \url{https://github.com/LuckyyySTA/GOLF}
\end{abstract}
\section{Introduction}
Recent advances in large language models (LLMs) have brought impressive progress in both preference alignment and reasoning. Achieving these capabilities typically relies on reinforcement learning (RL) methods, notably Reinforcement Learning with Human Feedback (RLHF)~\citep{ouyang2022training} and Reinforcement Learning with Verifiable Rewards (RLVR)~\citep{lambert2024tulu3, guo2025deepseekr1}, where training is driven exclusively by scalar outcome signals from reward models or automatic verifiers.

\begin{figure}[t]
    \centering
    \includegraphics[width=0.48\textwidth]{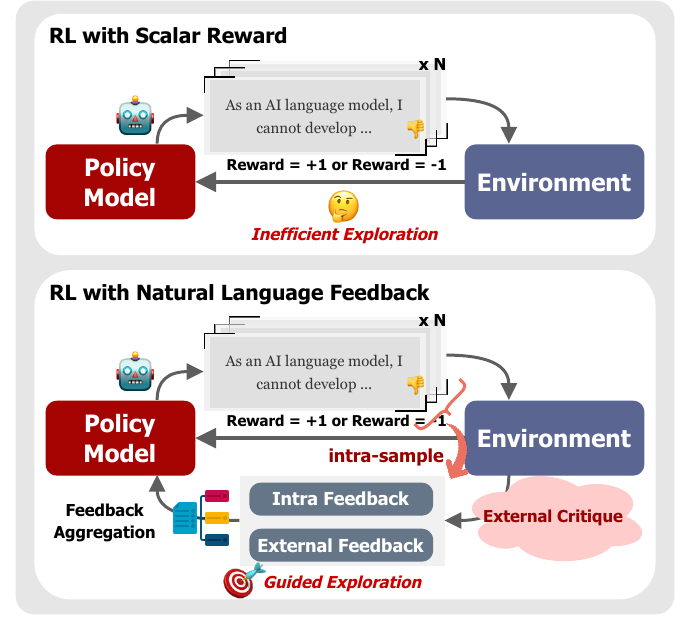}
    \caption{An illustration of RL with natural language feedabck. Compared with scalar-reward RL (top), aggregating intra-group and external feedback turns sparse outcomes into actionable refinement signals, enabling guided exploration (bottom).}
    \label{fig:overview}
\end{figure}
However, in many real-world scenarios, the supervision available to LLMs goes beyond sparse outcome rewards and often appears as natural language (NL) feedback, \eg, user feedback in human-model interactions~\citep{liu2025user} or textual judgments from generative reward models~\citep{ankner2024critique}.
Unlike scalar rewards, such feedback may take the form of explicit error diagnoses, comparisons across attempts, or concrete revision suggestions, offering more direct guidance for policy improvement.
Yet current RL algorithms, exemplified by GRPO~\citep{shao2024deepseekmath}, are not designed to fully exploit this richer supervision.
This limitation leads to \textit{inefficient exploration}: with only sparse outcome signals that indicate success or failure, the policy lacks explicit guidance on how to improve, forcing it to rely on costly trial and error to discover rewarding trajectories.
The issue is exacerbated when group-normalized advantages collapse (\eg, all-zero groups), yielding vanishing gradients~\citep{yu2025dapo} that halt learning entirely.

A promising avenue to address this limitation is to incorporate NL feedback into RL training~\citep{zhang2025critique, li2025lanpo}, translating it into actionable refinements that serve as explicit guidance to drive exploration.
In this work, we systematically exploit \textit{group-level} NL feedback from two 
complementary sources, as illustrated in Figure~\ref{fig:overview}: \emph{external critiques} that identify specific errors and suggest targeted revisions, and \emph{intra-group comparisons} derived from alternative attempts within the same rollout group, which reveal complementary partial ideas as well as diverse failure patterns. Our preliminary study in \S\ref{appendix:preliminary} demonstrates that aggregating both sources yields richer refinement contexts and produces more diverse, higher-quality refinements than either source alone.

Building on this insight, we propose \ours, an RL framework that explicitly leverages group-level NL feedback to improve exploration efficiency.
At its core, \ours consists of three tightly coupled components.
(1) \textbf{Aggregated Feedback Refinement}: We aggregate both external critiques and intra-group comparisons to produce refined responses for failed attempts. By combining these complementary sources, the resulting refinements not only correct identified errors but also explore diverse reasoning paths, broadening the policy's solution coverage.
(2) \textbf{Adaptive Refinement Injection}: When the current policy struggles to improve in sparse-reward regions, we \textit{adaptively} inject high-quality refinements as off-policy scaffolds to alleviate the exploration bottleneck. These scaffolds provide targeted guidance while maintaining the policy's exploration capacity.
(3) \textbf{Joint Optimization of Generation and Refinement}: We jointly optimize generation and refinement within a unified RL loop, so that improvements in self-refinement continuously raise the quality of the injected scaffolds, which in turn further improve exploration. Together, these components form a virtuous cycle between refinement quality and exploration efficiency.

To validate the effectiveness of \ours, we conduct extensive experiments on both verifiable and non-verifiable tasks, covering various model families and sizes.
Across five non-verifiable benchmarks, \ours achieves the best final performance, outperforming the strongest baseline by 22.7\%. Remarkably, \ours{} achieves approximately a 2.2$\times$ improvement in sample efficiency compared to vanilla RL methods trained solely on scalar rewards, significantly enhancing exploration efficiency.
On verifiable tasks, \ours yields consistent gains on math reasoning, instruction following, and code generation benchmarks, and further improves Pass@k, indicating broader solution coverage and diversity. Further analysis confirms that the gains are driven by complementary feedback sources and adaptive guidance, with each component contributing to the overall improvements.

Our contributions are summarized as follows:
\begin{itemize}
    \item We propose \ours, a novel RL framework that effectively aggregates group-level natural language feedback to guide exploration, substantially improving exploration efficiency.
    \item Extensive experiments on both verifiable and non-verifiable tasks show that external critiques and intra-group attempts provide complementary supervision, and that jointly integrating them into RL training yields superior performance.
    \item Comprehensive analysis reveals that \ours effectively promotes diverse exploration, maintaining higher training entropy and Pass@k performance, while simultaneously developing superior self-refinement capabilities by leveraging external feedback at inference time.
\end{itemize}
\section{Related Work}
\paragraph{Optimization with Natural Language Feedback.}
The idea of leveraging natural language feedback to improve model performance has been well explored. One line of work focuses on inference time optimization, where LLMs use textual feedback for self improvement by transforming it into self-reflective experiences~\citep{shinn2023reflexion} or by iteratively refining previous attempts~\citep{madaan2023selfrefine, kamoi2024when}. Another line of research learns from natural language feedback via imitation learning. For example, \citet{chen2024learning} fine-tunes models on high quality refinements through textual feedback, while \citet{wang2025critique} directly trains models to imitate critiques. Recent work also explores incorporating natural language feedback into the RL process. \citet{wang2025text2grad} and \citet{cao2024enhancing} train reward models that convert textual critiques into token level or span level reward signals, enabling finer grained credit assignment during reinforcement learning. Most similar to our work, Critique-GRPO~\citep{zhang2025critique} integrates critique guided refinements into online RL optimization. In contrast, we go beyond using only external critiques by additionally exploiting intra group feedback from multiple attempts, which provides richer signals about diverse failure modes and effectively extends to non-verifiable tasks.

\paragraph{Guiding Model Exploration in Reinforcement Learning.}
Exploration plays a critical role in reinforcement learning for LLMs. Yet LLMs' exploration capabilities remains bounded by the intrinsic capabilities of the current policy~\citep{yue2025does}, which may struggle to reach rewarding trajectories in low reward regimes. Recent work therefore introduces external supervision to guide exploration beyond the model’s default sampling distribution. LUFFY~\citep{yan2025learning} incorporates expert demonstrations as off policy samples to provide stronger learning signals. Critique-GRPO~\citep{zhang2025critique} leverages natural language feedback to identify errors and refine failed responses, using the resulting refinements as guided trajectories to facilitate exploration.
In contrast, our approach jointly trains problem solving and self refinement under outcome rewards, and uses the learned refinement capability to generate high quality refinement samples that serve as adaptive guidance for exploration, alleviating entropy collapse and accelerating the discovery of rewarding trajectories.

\section{Preliminaries}
Group Relative Policy Optimization (GRPO)~\citep{shao2024deepseekmath} is a widely used RL algorithm for training LLMs. It simplifies Proximal Policy Optimization (PPO)~\citep{schulman2017proximal} by eliminating the need for a trainable value function. For each prompt $x \in \mathcal{D}$, GRPO samples a group of $N$ responses $\{y^{(i)}\}_{i=1}^{N}$ with $y^{(i)}\sim\pi_{\theta_{\text{old}}}(\cdot\mid x)$.
A reward function assigns a scalar reward $r^{(i)}$ to each response. 
The group-relative advantage is computed by normalizing rewards within the group:
{
\begin{equation}
A^{(i)}
=
\frac{
r^{(i)} - \mathrm{mean}\!\left(\{r^{(j)}\}_{j=1}^{N}\right)
}{
\mathrm{std}\!\left(\{r^{(j)}\}_{j=1}^{N}\right)
}.
\label{eq:grpo_adv}
\end{equation}
}
GRPO then optimizes the clipped surrogate objective:
{
\begin{equation}
\begin{aligned}
\mathcal{J}_{\text{GRPO}}(\theta; x)
&=
\frac{1}{N}\sum_{i=1}^{N}\frac{1}{|y^{(i)}|}
\sum_{t=1}^{|y^{(i)}|}
\mathrm{CLIP}\Big(\rho^{(i)}_{t}(\theta),\,A^{(i)},\,\varepsilon\Big)\\
&\quad-\beta\,D_{\text{KL}}\!\left(\pi_{\theta}\,\|\,\pi_{\text{ref}}\right),
\end{aligned}
\label{eq:grpo_obj}
\end{equation}
}
{
\begin{equation}
\mathrm{CLIP}(r, A, \varepsilon)=\min\big(rA,\;\mathrm{clip}(r,\,1-\varepsilon,\,1+\varepsilon)\,A\big),
\label{eq:clip_def}
\end{equation}
}

where $\varepsilon$ and $\beta$ are hyperparameters controlling the clipping range and the KL penalty, respectively, and $\rho^{(i)}_{t}(\theta)= \frac{\pi_{\theta}(y^{(i)}_{t}\mid x, y^{(i)}_{<t})} {\pi_{\theta_{\text{old}}}(y^{(i)}_{t}\mid x, y^{(i)}_{<t})}$ is the importance sampling ratio.
Following prior work~\citep{yan2025learning}, we adopt the Dr.\ GRPO~\citep{liu2025understanding} variant by removing length normalization and standard-deviation normalization throughout this paper.

\section{Methodology}
\begin{figure*}[h]
    \centering
    \includegraphics[width=1.0\textwidth]{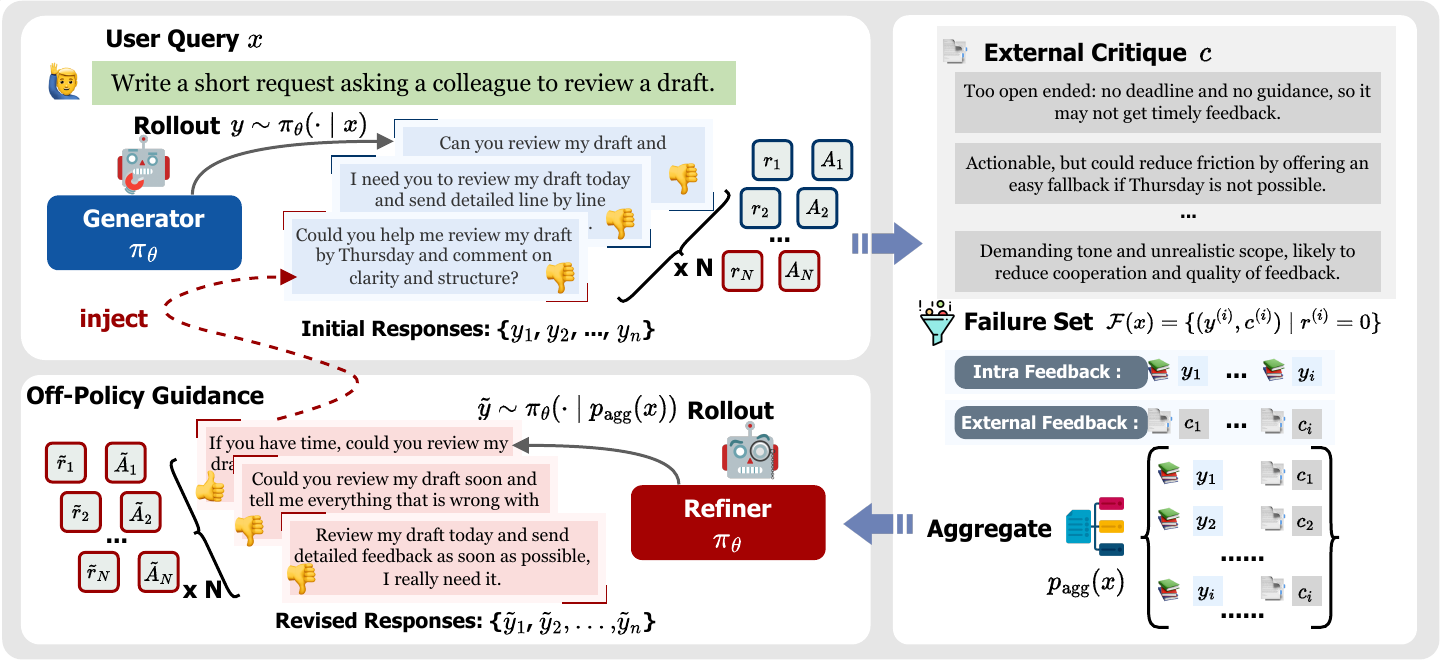}
    \caption{An overview of \ours, which consists of three components. The policy first rollouts a group of candidates and receives both scalar rewards and external critiques. \ours then aggregates the critiques with the failed trajectories in the same group to form group-level NL feedback, which conditions a refinement stage to produce improved responses. Finally, high-quality refinements are adaptively injected back into the rollout group as off-policy guidance, mitigating low-reward regimes. Both generation and refinement are optimized jointly within a unified RL loop.}
    \label{fig:framework}
\end{figure*}
In this section, we describe \ours, which consists of three tightly coupled components, as illustrated in Figure~\ref{fig:framework}.
\subsection{Group-level Feedback Aggregated Refinement}
\label{sec:agg_refine}

For each prompt $x \sim \mathcal{D}$, we sample a group of $N$ responses
$\mathcal{G}_{\text{gen}}(x)=\{y^{(i)}\}_{i=1}^{N}$, and query the reward model for the scalar reward and corresponding critique:
$
(r^{(i)}, c^{(i)}) = R(x, y^{(i)}).
$
We consider two types of NL feedback: (1) \textbf{External feedback} refers to the critique $c^{(i)}$ associated with a specific response $y^{(i)}$; (2) \textbf{Intra-group feedback} refers to alternative responses in the group $\mathcal{G}_{\text{gen}}(x)$, which often contain complementary partial ideas.

Instead of refining each failure in isolation, we \textit{aggregate} multiple failed attempts in the group together with their critiques into a single refinement context, exposing diverse failure modes.
Concretely, we collect the failure set
$
\mathcal{F}(x)=\{(y^{(i)}, c^{(i)}) \mid r^{(i)}=0\},
$
and construct the aggregated refinement prompt (see Appendix~\ref{appendix:refine_prompts} for prompts):
\begin{equation}
p_{\text{agg}}(x)
=
\textsc{Concat}\Bigl(x,\; \mathcal{F}(x)\Bigr).
\label{eq:agg_prompt}
\end{equation}
Conditioned on $p_{\text{agg}}(x)$, we sample a refinement group $\mathcal{G}_{\text{refine}}(x)=\{\tilde y^{(j)}\}_{j=1}^{N}$ with
$\tilde y^{(j)}\sim \pi_{\theta_{\text{old}}}(\cdot \mid p_{\text{agg}}(x))$ and score each refinement by
$\tilde r^{(j)} = R(x, \tilde y^{(j)})$.
This aggregation enables synthesis that identifies mistakes, fills gaps, and composes complementary partial ideas, thereby producing refinements that surpass any single attempt.

\subsection{Adaptive Guidance via Mixed Policy Optimization}
\label{sec:mixed}

In low reward regimes, on-policy groups often contain only zero-reward samples, yielding weak group-relative advantages and slow policy improvement. We therefore treat high quality refinements as off policy scaffolds and inject them into the sampled group to restore informative advantages, while optimizing the policy with a mixed objective that combines on policy and off policy trajectories.

\paragraph{Adaptive Injection.}
For each prompt $x$, we first compute the average reward in the group $\mathcal{G}_{\text{gen}}(x)$:
\begin{equation}
s(x)=\frac{1}{N}\sum_{y\in \mathcal{G}_{\text{gen}}(x)} r(x,y).
\end{equation}
We trigger injection only in low-reward regimes, \eg, when the group's average reward $s(x)$ falls below a threshold $\tau$, set to $1/N$ by default.
In that case, we form the set of \emph{successful} refinements:
\begin{equation}
\mathcal{S}_{\text{ref}}(x)=\{\tilde y \in \mathcal{G}_{\text{ref}}(x)\mid \tilde r(x,\tilde y)=1\}.
\end{equation}
If $\mathcal{S}_{\text{ref}}(x)\neq\emptyset$, we randomly select a $\tilde y^\star \in \mathcal{S}_{\text{ref}}(x)$ and inject it by randomly replacing one failed response in $\mathcal{G}_{\text{gen}}(x)$.

\paragraph{Mixed Policy Optimization.}
Let $\mathcal{G}_{\text{aug}}(x)=\mathcal{G}_{\text{on}}(x)\cup\mathcal{G}_{\text{off}}(x)$, where
$\mathcal{G}_{\text{on}}(x)$ are rollouts sampled from $\pi_{\theta_{\text{old}}}(\cdot\mid x)$ and
$\mathcal{G}_{\text{off}}(x)$ are injected refinement trajectories generated under the refinement prompt
$p_{\text{agg}}(x)$ by the same policy $\pi_{\theta_{\text{old}}}$. 
Then, we utilize the following mixed policy optimization objective~\citep{yan2025learning} to update the policy $\pi_\theta$:
{\small
\begin{equation}
\begin{aligned}
\mathcal{J}_{\text{Mixed}}(\theta)
&=
\frac{1}{Z}\Bigg[
\underbrace{
\sum_{i=1}^{N_{\text{on}}}\sum_{t=1}^{|\tau_i|}
\mathrm{CLIP}\Big(r^{\text{on}}_{i,t}(\theta),\hat A_i,\varepsilon\Big)
}_{\text{on-policy objective}}
\\[4pt]
&\quad+
\underbrace{
\sum_{j=1}^{N_{\text{off}}}\sum_{t=1}^{|\tau_j|}
\mathrm{CLIP}\Big(f(r^{\text{off}}_{j,t}(\theta)),\hat A_j,\varepsilon\Big)
}_{\text{off-policy objective}}
\Bigg],
\end{aligned}
\label{eq:mixed_obj}
\end{equation}
}

where $Z\;=\; \sum_{i=1}^{N_{\text{on}}} |\tau_i| \;+\; \sum_{j=1}^{N_{\text{off}}} |\tau_j| $ normalizes by the total number of tokens, and
{
\begin{equation}
\begin{aligned}
r^{\text{on}}_{i,t}(\theta)
&=
\frac{\pi_\theta(\tau_{i,t}\mid x,\tau_{i,<t})}
{\pi_{\theta_{\text{old}}}(\tau_{i,t}\mid x,\tau_{i,<t})},
\\[4pt]
r^{\text{off}}_{j,t}(\theta)
&=
\frac{\pi_\theta(\tau_{j,t}\mid x,\tau_{j,<t})}
{\pi_{\theta_{\text{old}}}(\tau_{j,t}\mid p_{\text{agg}}(x),\tau_{j,<t})}.
\end{aligned}
\label{eq:ratios}
\end{equation}
}

We compute advantages by normalizing rewards within the augmented group $\mathcal{G}_{\text{aug}}(x) = \mathcal{G}_{\text{on}}(x)\cup \mathcal{G}_{\text{off}}(x)$:
{
\begin{equation}
\hat A_i
=
R(\tau_i)-\mathrm{mean}\Big(\mathcal{G}_{\text{aug}}(x)\Big).
\label{eq:adv_mixed}
\end{equation}
}

Following prior work~\citep{yan2025learning}, we apply the reshaping function
$f(u)=u/(u+\lambda)$ with $\lambda=0.1$ to off-policy ratios, and omit the clip operation for off-policy rollouts to emphasize low-probability yet effective actions from injected refinements.

\subsection{Joint Optimization for Self-Refinement}
\label{sec:joint}
During post-training, LLMs are trained with RL to improve problem solving ability, while test-time self-refinement is not explicitly accounted for. Empirically, we observe that standard RL fine-tuning can even degrade performance when combined with test-time self-refinement. 
To address this, we explicitly train the LLM to improve both direct problem solving and feedback-conditioned refinement within one integrated RL process.

Concretely, for each prompt $x$, we collect two rollout groups: a generation group
$\mathcal{G}_{\text{gen}}(x)$ and a refinement group
$\mathcal{G}_{\text{ref}}(x)$. We then concatenate them into a joint batch
$\mathcal{B}(x)=\mathcal{G}_{\text{gen}}(x)\cup \mathcal{G}_{\text{ref}}(x)$.
The advantages within each group are computed separately, and then update the policy $\pi_\theta$ using GRPO within a single RL process.

Jointly optimizing these two behaviors forms a positive feedback loop: as self-refinement improves, it produces higher-quality refinement trajectories that serve as stronger off-policy scaffolds for mixed-policy optimization, increasing the likelihood of discovering rewarding trajectories.
\section{\ours on Non-verifiable Tasks}
\label{sec:fuzzy}
\definecolor{steelbluev2}{HTML}{DAE8FC}
\definecolor{steelblue}{HTML}{82B0D2}
\definecolor{mygray}{HTML}{808080}

\begin{table*}[h]
\centering
\caption{Experimental results on non-verifiable tasks. All metrics are the higher the better. \textbf{Bold} and \ul{underline} numbers indicate the best performance and second performance among all methods, respectively. WildBench scores are in $[-100,100]$, while all other metrics are in $[0,100]$. All scores are judged by GPT-4o.}
\label{tab:fuzzy_main_results}
\resizebox{\textwidth}{!}{%
\begin{tabular}{lccccccc}
\toprule
\multirow{3}{*}{\textbf{Model}}
& \multicolumn{2}{c}{\textbf{AlpacaEval-v2}}
& \multicolumn{1}{c}{\textbf{WildBench}}
& \multicolumn{1}{c}{\textbf{Arena-Hard-v1}}
& \multicolumn{1}{c}{\textbf{ArenaHard-v2}}
& \multicolumn{1}{c}{\textbf{CreativeWriting-v3}}
& \multicolumn{1}{c}{\textbf{Average}} \\
\cmidrule(lr){2-3}
\cmidrule(lr){4-4}
\cmidrule(lr){5-5}
\cmidrule(lr){6-6}
\cmidrule(lr){7-7}
\cmidrule(lr){8-8}
& \textbf{Win Rate (\%)} & \textbf{LC Win Rate (\%)}
& \textbf{LLM Judge (\%)}
& \textbf{Win Rate (\%)}
& \textbf{Win Rate (\%)}
& \textbf{LLM Judge (\%)}
& \textbf{Score (\%)} \\
\midrule
\textbf{\textit{Llama-3.1-8B-Instruct}}
& 31.93 & 31.79
& -8.25
& 30.80
& 5.57
& 53.96
& 24.30 \\
+ Direct-Likert
& 38.88 & 34.98
& 13.48
& 51.55
& 11.73
& 64.10
& 35.79 \\
+ Pairwise-GRPO
& 45.47 & 43.19
& 25.54
& 49.20
& 13.30
& 62.95
& 39.94 \\
+ Rubric-as-Reward
& 42.24 & 36.12
& \ul{26.51}
& \ul{52.10}
& \ul{15.57}
& \textbf{68.12}
& 40.11 \\
+ Critique-GRPO
& \ul{47.45} & \ul{43.31}
& 25.09
& 50.15
& 13.73
& 65.76
& \ul{40.92} \\
\rowcolor{steelblue!33}
+ \ours
& \textbf{53.42} & \textbf{69.67}
& \textbf{34.42}
& \textbf{52.40}
& \textbf{25.03}
& \ul{66.21}
& \textbf{50.19} \\
\midrule
\textbf{\textit{Qwen-3-8B}}
& 55.16 & 52.60
& 48.05
& 70.70
& 33.90
& 63.27
& 53.95 \\
+ Direct-Likert
& 64.84 & 61.06
& 58.01
& \textbf{82.75}
& 41.70
& \ul{69.56}
& 62.99 \\
+ Pairwise-GRPO
& 66.34 & 68.34
& \ul{67.77}
& 81.20
& \ul{50.10}
& 68.08
& 66.97 \\
+ Rubric-as-Reward
& 65.34 & 68.88
& 67.09
& 81.90
& 50.08
& 69.21
& \ul{67.08} \\
+ Critique-GRPO
& \ul{68.20} & \ul{69.82}
& 64.84
& \ul{81.95}
& 49.63
& 67.30
& 66.96 \\
\rowcolor{steelblue!33}
+ \ours
& \textbf{71.80} & \textbf{71.94}
& \textbf{68.16}
& 80.90
& \textbf{52.00}
& \textbf{70.78}
& \textbf{69.26} \\
\bottomrule
\end{tabular}
}
\end{table*}

In this section, we demonstrate the effectiveness of \ours on general non-verifiable tasks. We first describe the setup in \S\ref{ssec:fuzzy-setup} and then present the main results in \S\ref{ssec:fuzzy-exps}.

\subsection{Experimental Setup}
\label{ssec:fuzzy-setup}

\paragraph{Models and Training Data.}
We experiment with two model families: \texttt{Llama-3.1-8B-Instruct} and \texttt{Qwen-3-8B} in non-thinking mode.
Following prior work~\citep{bhaskar2025language}, we train on 7{,}500 prompts from \textit{WildChat-IF}, a subset of WildChat~\citep{zhao2024wildchat} that prioritizes conversational instructions.

\paragraph{Baselines.}
We compare our method against the following baselines, all built on GRPO: (1) \textbf{\textit{Direct-Likert}}, where an LLM-as-judge provides a direct assessment for each prompt-response pair on a 1--10 Likert scale. The resulting score is used as the reward for RL training; (2) \textbf{\textit{Pairwise-GRPO}}, where a binary reward is computed through pairwise comparisons against a high-quality reference response generated by GPT-4o; (3) \textbf{\textit{Rubric-as-Reward}}~\citep{gunjal2025rubrics}, where a judge model evaluates each response using prompt-specific rubrics generated by DeepSeek-v3.2~\citep{deepseekai2025deepseekv32} and assigns a single 1--10 Likert rating; (4) \textbf{\textit{Critique-GRPO}}~\citep{zhang2025critique}, which solely leverages external critiques to guide policy refinement while using pairwise comparison-based reward to steer RL training.

\paragraph{Evaluation Benchmarks.}
We evaluate on five standard non-verifiable benchmarks spanning general chat, instruction following, and creative writing: AlpacaEval v2.0~\citep{alpaca_eval, dubois2024length}, ArenaHard v1.0~\citep{arenahard2024}, ArenaHard v2.0~\citep{li2024crowdsourced}, WildBench~\citep{lin2025wildbench}, and CreativeWritingV3~\citep{creative-writing-bench-v3}.
We report win rate (WR) and length-controlled win rate (LC-WR) on AlpacaEval v2.0, and WR on both ArenaHard v1.0/v2.0.
Following \citet{bhaskar2025language}, we use an LLM-as-a-judge to compute the WildBench score in $[-100,100]$ and the CreativeWritingV3 score in $[0,100]$.
Detailed benchmark descriptions are provided in Appendix~\ref{appendix:benchmarks}.

\paragraph{Experimental Details.}
During RL training, we use \texttt{Qwen3-235B-A22B-Instruct-2507} as the judge to produce both scalar rewards and external critiques. We train for 2 epochs for all methods, and report the best performance.
For evaluation, following \citet{li2025jointly} and \citet{bhaskar2025language}, we use GPT-4o as the judge across all benchmarks.
Additional experimental details are provided in Appendix~\ref{appendix:fuzzy_training_details}.

\subsection{Experimental Results}
\label{ssec:fuzzy-exps}

\paragraph{\ours achieves the best performance across non-verifiable benchmarks.}
Table~\ref{tab:fuzzy_main_results} reports the main results on five non-verifiable benchmarks.
Across both model families, \ours achieves the best average score, surpassing the strongest baseline by \textbf{+9.27} points on \texttt{Llama-3.1-8B-Instruct} (50.19 vs.\ 40.92) and by \textbf{+2.18} points on \texttt{Qwen-3-8B} (69.26 vs.\ 67.08).
Moreover, compared with Critique-GRPO, which relies solely on external critiques, \ours further improves the average by \textbf{+9.27} and \textbf{+2.30} points on the two backbones, respectively.
These results indicate that incorporating group-level feedback into online reinforcement learning effectively strengthens the model's general capabilities.

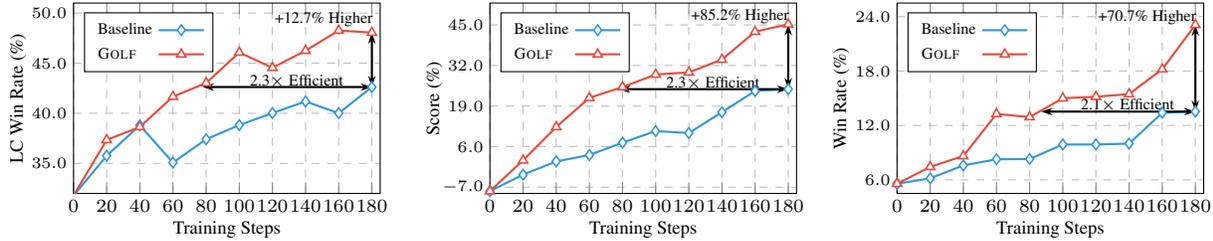
\begin{figure*}[t]
    \centering
    \definecolor{upurple}{RGB}{155,89,182}
\definecolor{ublue}{RGB}{52,152,219}
\definecolor{ured}{RGB}{231,76,60}
\definecolor{udark}{RGB}{77,153,77}
\definecolor{ugreen}{RGB}{46,204,113}
\begin{tikzpicture}
\scriptsize{
\begin{axis}[
at={(0,0)},
width=.33\textwidth, height=.24\textwidth ,
xtick={0,20,40,60,80,100,120,140,160,180},  % 只显示这些刻度
   % xticklabels={0,20,40,60,80,100,120,140,160,180},  % 刻度标签
ytick={35,40,45,50,55,60,65},
xmin=0,  % 添加这一行，让x轴从-10开始，给0留出显示空间
xmax=185,  % 添加这一行，让x轴从-10开始，给0留出显示空间
ymin=32,ymax=51,
xlabel={Training Steps},
xlabel style={yshift=1em,},
grid style=dashed,
ylabel={LC Win Rate (\%)},
xlabel style={align=center,font=\scriptsize},
ylabel style={font=\scriptsize,yshift=-2em},
y tick style={opacity=0},
y tick label style={font=\tiny},
ymajorgrids=true,
xmajorgrids=true,
tick align=inside,
legend pos=outer north east,
yticklabel style={/pgf/number format/precision=1,/pgf/number format/fixed zerofill},
legend style={yshift=-0.5em,xshift=-16.5em,legend cell align=left,legend plot pos=right},]

\addplot [sharp plot,ublue,thick,line width=0.5pt,mark=diamond*,mark size=2pt,thick,mark options={fill=white,draw=ublue,line width=0.6pt}] plot coordinates {
    (0,31.79)
      (20,35.76)(40,38.80) (60,35.08)
      (80,37.41) (100,38.82) (120,40.02)
      (140,41.18) (160,40.01) (180,42.62)
      % (200,43.19) 
      };
      % ours
\addplot [sharp plot,ured,thick,line width=0.5pt,mark=triangle*,,mark size=2pt,thick,mark options={fill=white,draw=ured,line width=0.6pt}] plot coordinates {
(0,31.79)
      (20,37.34)(40,38.64) (60,41.66)
      (80,43.02) (100,46.07) (120,44.54)
      (140,46.25) (160,48.26) (180,48.05)
      % (200,45.43) 
      };
\draw[{Stealth[length=1.5mm, width=1mm]}-{Stealth[length=1.5mm, width=1mm]},thick,black] 
    (axis cs:180,42.62) -- (axis cs:180,48.05)
node[midway,above,xshift=-2.6em,yshift=1.4em,font=\tiny,text=black] {+12.7\% Higher};
\draw[{Stealth[length=1.5mm, width=1mm]}-{Stealth[length=1.5mm, width=1mm]},thick,black] (axis cs:180,42.62) -- (axis cs:78,42.62)node[midway,above,font=\tiny,xshift=0.5em,yshift=-0.4em,text=black] {2.3$\times$ Efficient};

\legend{\tiny{Baseline},\tiny{\ours}},
\end{axis}
}
\vspace{6cm}

\scriptsize{
\begin{axis}[
at={(22.5em,0)},
width=.33\textwidth, height=.24\textwidth ,
xtick={0,20,40,60,80,100,120,140,160,180},  
ytick={-7.0,6.0,19.0,32.0,45.0},
xmin=0,  
xmax=185,  
ymin=-9,ymax=52,
xlabel={Training Steps},
xlabel style={yshift=1em,},
grid style=dashed,
ylabel={Score (\%)},
xlabel style={align=center,font=\scriptsize},
ylabel style={font=\scriptsize,yshift=-2em,},
y tick style={opacity=0},
y tick label style={font=\tiny},
ymajorgrids=true,
xmajorgrids=true,
tick align=inside,
legend pos=outer north east,
legend style={yshift=-0.5em,xshift=-16.5em,legend cell align=left,legend plot pos=right},
yticklabel style={/pgf/number format/precision=1,/pgf/number format/fixed zerofill},]

% baseline
\addplot [sharp plot,ublue,thick,line width=0.5pt,mark=diamond*,mark size=2pt,thick,mark options={fill=white,draw=ublue,line width=0.6pt}] plot coordinates {
    (0,-8.25)
      (20,-2.98)(40,1.27) (60,3.32)
      (80,7.28) (100,10.99) (120,10.35)
      (140,17.04) (160,23.83) (180,24.41)
      % (200,43.19) 
      };
      % ours
\addplot [sharp plot,ured,thick,line width=0.5pt,mark=triangle*,,mark size=2pt,thick,mark options={fill=white,draw=ured,line width=0.6pt}] plot coordinates {
(0,-8.25)
      (20,1.61)(40,12.3) (60,21.63)
      (80,25.1) (100,29.1) (120,29.79)
      (140,33.89) (160,42.82) (180,45.21)
      % (200,45.43) 
      };

\draw[{Stealth[length=1.5mm, width=1mm]}-{Stealth[length=1.5mm, width=1mm]},thick,black] 
    (axis cs:180,24.41) -- (axis cs:180,45.21)
node[midway,above,xshift=-2.6em,yshift=1.4em,font=\tiny,text=black] {+85.2\% Higher};
\draw[{Stealth[length=1.5mm, width=1mm]}-{Stealth[length=1.5mm, width=1mm]},thick,black] (axis cs:180,24.41) -- (axis cs:78,24.41)node[midway,above,font=\tiny,xshift=0.5em,yshift=-0.4em,text=black] {2.3$\times$ Efficient};

\legend{\tiny{Baseline},\tiny{\ours}},
\end{axis}
}

\scriptsize{
\begin{axis}[
at={(44.5em,0)},
width=.33\textwidth, height=.24\textwidth ,
xtick={0,20,40,60,80,100,120,140,160,180}, 
ytick={6,12,18,24},
xmin=0,  
xmax=185,  
ymin=4.5,ymax=25.5,
xlabel={Training Steps},
xlabel style={yshift=1em,},
grid style=dashed,
ylabel={Win Rate (\%)},
xlabel style={align=center,font=\scriptsize},
ylabel style={font=\scriptsize,yshift=-2em},
y tick style={opacity=0},
y tick label style={font=\tiny},
ymajorgrids=true,
xmajorgrids=true,
tick align=inside,
legend pos=outer north east,
legend style={yshift=-0.5em,xshift=-16.5em,legend cell align=left,legend plot pos=right},
yticklabel style={/pgf/number format/precision=1,/pgf/number format/fixed zerofill},]

% baseline
\addplot [sharp plot,ublue,thick,line width=0.5pt,mark=diamond*,mark size=2pt,thick,mark options={fill=white,draw=ublue,line width=0.6pt}] plot coordinates {
    (0,5.57)
      (20,6.17)(40,7.6) (60,8.27)
      (80,8.3) (100,9.9) (120,9.9)
      (140,10) (160,13.4) (180,13.53)
      % (200,43.19) 
      };
      % ours
\addplot [sharp plot,ured,thick,line width=0.5pt,mark=triangle*,,mark size=2pt,thick,mark options={fill=white,draw=ured,line width=0.6pt}] plot coordinates {
(0,5.57)
      (20,7.43)(40,8.63) (60,13.27)
      (80,12.93) (100,15) (120,15.17)
      (140,15.46) (160,18.2) (180,23.1)
      % (200,45.43) 
      };

\draw[{Stealth[length=1.5mm, width=1mm]}-{Stealth[length=1.5mm, width=1mm]},thick,black] 
    (axis cs:180,13.53) -- (axis cs:180,23.1)
node[midway,above,xshift=-2.6em,yshift=2em,font=\tiny,text=black] {+70.7\% Higher};
\draw[{Stealth[length=1.5mm, width=1mm]}-{Stealth[length=1.5mm, width=1mm]},thick,black] (axis cs:180,13.53) -- (axis cs:87,13.53)node[midway,above,font=\tiny,xshift=0.5em,yshift=-0.4em,text=black] {2.1$\times$ Efficient};

\legend{\tiny{Baseline},\tiny{\ours}},
\end{axis}
}

\end{tikzpicture}
  \caption{Evaluation performance over training steps. We report the LC win rate on AlpacaEval v2.0 (left), WildBench score (middle), and ArenaHard v2.0 win rate (right). The baseline refers to Pairwise-GRPO, which uses the same generative reward model as \ours.}
  \label{fig:fuzzy_evaluation_over_steps}
\end{figure*}

\paragraph{\ours significantly improves both exploration efficiency and the performance ceiling.}
Figure~\ref{fig:fuzzy_evaluation_over_steps} tracks evaluation performance as training progresses.
Across all three benchmarks, \ours improves more rapidly in the early stage, reaching comparable performance with substantially fewer training steps.
For instance, on AlpacaEval v2.0, \ours matches the baseline final LC win rate in just 80 steps, yielding a 2.25$\times$ improvement in sample efficiency.
The same pattern holds on WildBench and ArenaHard v2.0, where \ours achieves 2.3$\times$ and 2.1$\times$ sample efficiency over the baseline, respectively.
Moreover, as training progresses, \ours ultimately converges to a consistently higher plateau, outperforming the baseline by \textbf{+12.7\%} on AlpacaEval v2.0, \textbf{+85.2\%} on WildBench, and \textbf{+70.7\%} on ArenaHard v2.0.
These results indicate that leveraging group-level natural language feedback both accelerates policy learning and raises the achievable performance ceiling.
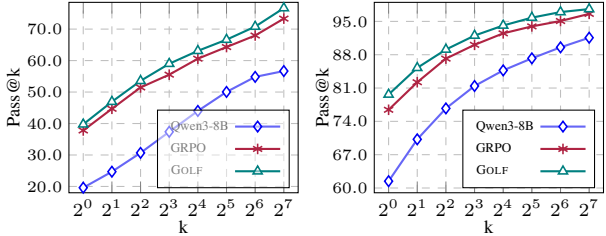
\begin{figure}[!t]
    \centering
    \definecolor{upurple}{RGB}{155,89,182}
\definecolor{ublue}{RGB}{52,152,219}
\definecolor{ured}{RGB}{231,76,60}
\definecolor{udark}{RGB}{77,153,77}
\definecolor{ugreen}{RGB}{46,204,113}
\definecolor{draw1}{RGB}{102, 102, 246}
\definecolor{fill1}{RGB}{0, 0, 245}
\definecolor{draw2}{RGB}{227, 182, 198}
\definecolor{fill2}{RGB}{175, 36, 67}
\begin{tikzpicture}

\scriptsize{
\begin{axis}[
at={(0em,0)},
width=.27\textwidth, height=.24\textwidth ,
xtick={0,1,2,3,4,5,6,7},
xticklabels={$2^0$,$2^1$,$2^2$,$2^3$,$2^4$,$2^5$,$2^6$,$2^7$},
xmin=-0.5,
xmax=7.5,
xlabel={$k=2^i$},
ymin=18,ymax=78.5,
ytick={20,30,40,50,60,70},
xlabel={k},
xlabel style={yshift=1em,},
grid style=dashed,
ylabel={Pass@k},
xlabel style={align=center,font=\scriptsize},
ylabel style={font=\scriptsize,yshift=-2em,},
y tick style={opacity=0},
%x tick label style={font=\small},
y tick label style={font=\tiny},
ymajorgrids=true,
xmajorgrids=true,
tick align=inside,
legend pos=outer north east,
legend style={yshift=-5.8em,xshift=-8em,legend cell align=left,legend plot pos=right,fill opacity=0.5,font={\fontsize{5}{0}\selectfont},/tikz/every even column/.append style={column sep=0.1cm},},
yticklabel style={/pgf/number format/precision=1,/pgf/number format/fixed zerofill},]

% Qwen3-8B
\addplot [sharp plot,draw1,thick,line width=0.5pt,mark=diamond*,mark size=2pt,thick,mark options={fill=white,draw=fill1,line width=0.6pt}] plot coordinates {
    (0,19.61)
      (1,24.69)(2,30.67) (3,37.37)
      (4,44) (5,50.04) (6,54.82)
      (7,56.67) 
      % (200,43.19) 
      };
% GRPO
\addplot [sharp plot,fill2,thick,line width=0.5pt,mark=asterisk,mark size=2pt,thick,mark options={fill=white,draw=fill2,line width=0.6pt}] plot coordinates {
    (0,37.79)
      (1,44.68)(2,51.46) (3,55.52)
      (4,60.6) (5,64.30) (6,67.96)
      (7,73.33) 
      };
% ours
\addplot [sharp plot,teal,thick,line width=0.5pt,mark=triangle*,,mark size=2pt,thick,mark options={fill=white,draw=teal,line width=0.6pt}] plot coordinates {
(0,39.74)
      (1,46.92)(2,53.55) (3,58.99)
      (4,63.09) (5,66.64) (6,70.79)
      (7,76.67) 
      };

\legend{Qwen3-8B,GRPO,\ours},
\end{axis}
}

\scriptsize{
\begin{axis}[
at={(16.5em,0)},
width=.27\textwidth, height=.24\textwidth ,
xtick={0,1,2,3,4,5,6,7},
xticklabels={$2^0$,$2^1$,$2^2$,$2^3$,$2^4$,$2^5$,$2^6$,$2^7$},
xmin=-0.5,
xmax=7.5,
xlabel={$k=2^i$},
ymin=59,ymax=99,
ytick={60,67,74,81,88,95},
xlabel={k},
xlabel style={yshift=1em,},
grid style=dashed,
ylabel={Pass@k},
xlabel style={align=center,font=\scriptsize},
ylabel style={font=\scriptsize,yshift=-2.2em},
y tick style={opacity=0},
y tick label style={font=\tiny},
ymajorgrids=true,
xmajorgrids=true,
tick align=inside,
legend pos=outer north east,
legend style={yshift=-5.8em,xshift=-8em,legend cell align=left,legend plot pos=right,font={\fontsize{5}{0}\selectfont},/tikz/every even column/.append style={column sep=0.1cm},},
yticklabel style={/pgf/number format/precision=1,/pgf/number format/fixed zerofill},]
\addplot [sharp plot,draw1,thick,line width=0.5pt,mark=diamond*,mark size=2pt,thick,mark options={fill=white,draw=fill1,line width=0.6pt}] plot coordinates {
    (0,61.43)
      (1,70.23)(2,76.72) (3,81.43)
      (4,84.71) (5,87.24) (6,89.55)
      (7,91.57) 
      };
% GRPO
\addplot [sharp plot,fill2,thick,line width=0.5pt,mark=asterisk,mark size=2pt,thick,mark options={fill=white,draw=fill2,line width=0.6pt}] plot coordinates {
    (0,76.43)
      (1,82.22)(2,87.18) (3,90.08)
      (4,92.49) (5,93.95) (6,95.08)
      (7,96.58) 
      };
% ours
\addplot [sharp plot,teal,thick,line width=0.5pt,mark=triangle*,,mark size=2pt,thick,mark options={fill=white,draw=teal,line width=0.6pt}] plot coordinates {
(0,79.64)
      (1,85.24)(2,89.09) (3,92.02)
      (4,94.14) (5,95.79) (6,96.96)
      (7,97.59) 
      };

\legend{Qwen3-8B,GRPO,\ours},
\end{axis}
}

\end{tikzpicture}
  \caption{Pass@$k$ comparison between GRPO and \ours on mathematical reasoning benchmarks using Qwen-3-8B.}
  \label{fig:pass_at_k}
\end{figure}

\section{\ours on Verifiable Tasks}
This section presents the main results of \ours on verifiable tasks, covering mathematical reasoning, instruction following, and code generation. We describe the setup in \S\ref{ssec:verifiable-setup} and present the main results in \S\ref{ssec:verifiable-exp}.

\subsection{Experimental Setup}
\label{ssec:verifiable-setup}
\paragraph{Models and Training Data.}
We conduct experiments on the Qwen3~\citep{yang2025qwen3} model family at two scales: \texttt{Qwen-3-4B} and \texttt{Qwen-3-8B}, using the non-thinking mode.
For training data, we follow the prior study~\citep{zhang2025critique}, using a high-quality subset from OpenR1-Math~\citep{openr1math} as our mathematical reasoning training set, comprising 4,000 problems. For instruction-following tasks, we utilize the IFTrain training data provided by \citet{pyatkin2025generalizing}, where we further filter out instructions that are unanswerable due to conflicting constraints or low quality, resulting in 3,798 high-quality samples.
For code generation, we adopt the LCBv6 subset of LiveCodeBench~\citep{jain2025livecodebench}, comprising competitive programming problems at varying difficulty levels. Following the concurrent work~\citep{hübotter2026reinforcementlearningselfdistillation}, we randomly sample 50\% of private tests as public tests for training, while the remaining are held out for evaluation.

\paragraph{Baselines.}
For mathematical reasoning and instruction following, we compare against both supervised fine-tuning and RL methods trained on the same datasets: (1) \textbf{Refinement-FT}~\citep{chen2024learning}, which fine-tunes the model with \textit{best-of-n} refinements conditioned on initial responses and external critiques; (2) \textbf{Critique-FT}~\citep{wang2025critique}, which fine-tunes the model to imitate high-quality critiques; (3) \textbf{GRPO}~\citep{shao2024deepseekmath}, group relative policy optimization with binary outcome rewards verified against ground-truth answers; (4) \textbf{Critique-GRPO}~\citep{zhang2025critique}, which integrates a critique-guided refinement mechanism into the RL loop.
For code generation, we follow the experimental setup of the concurrent work SDPO~\citep{hübotter2026reinforcementlearningselfdistillation} and compare against a strong GRPO baseline with $\varepsilon_{\text{high}}=0.28$~\citep{yu2025dapo}, as well as \textbf{SDPO} itself, which utilizes execution feedback as hindsight information for on-policy distillation.

\definecolor{steelbluev2}{HTML}{DAE8FC}
\definecolor{steelblue}{HTML}{82B0D2}
\definecolor{mygray}{HTML}{808080}

\begin{table}[t]
\centering
\caption{Experimental results on verifiable tasks. All metrics are higher is better. \textbf{Bold} and \ul{underline} numbers indicate the best performance and second performance among all methods.}
\label{tab:verifiable_main_results}
\resizebox{\columnwidth}{!}{%
% \begin{tabular}{llllll}
\begin{tabular}{lccccc}
\toprule
\multirow{3}{*}{\textbf{Model}}
& \multicolumn{3}{c}{\textbf{Mathmatical Reasoning}}
& \multicolumn{2}{c}{\textbf{Instruction Following}} \\
\cmidrule(lr){2-4}
\cmidrule(lr){5-6}
& \multicolumn{1}{c}{\textbf{AIME 24}}
& \multicolumn{1}{c}{\textbf{AIME 25}}
& \multicolumn{1}{c}{\textbf{AMC 23}}
& \multicolumn{1}{c}{\textbf{IFBench}}
& \multicolumn{1}{c}{\textbf{IFEval}} \\
\midrule
\textbf{\textit{Qwen-3-4B}}
& 22.53 & 18.55 & 59.41 & 23.67 & 81.52 \\
+ Refinement-FT
& 31.67 & 21.25 & 64.06 & 30.44 & 83.73 \\
+ Critique-FT
& 34.58 & 24.58 & 65.94 & 31.67 & 82.63 \\
+ GRPO
& 42.72 & 35.42 & \ul{76.85} & 33.33 & 84.45 \\
+ Critique-GRPO
& \ul{45.72} & \ul{35.89} & 76.14 & \ul{35.67} & \ul{85.21} \\
\rowcolor{steelblue!33}
+ \ours
& \textbf{49.18} & \textbf{38.10} & \textbf{77.15} & \textbf{37.67} & \textbf{86.51} \\
\midrule
\textbf{\textit{Qwen-3-8B}}
& 27.97 & 19.60 & 61.32 & 27.00 & 83.55 \\
+ Refinement-FT
& 42.08 & 27.50 & 67.81 & 34.33 & 84.29 \\
+ Critique-FT
& 46.75 & 28.75 & 70.31 & 33.60 & 84.45 \\
+ GRPO
& 55.05 & \ul{38.02} & \ul{78.61} & 35.65 & 84.76 \\
+ Critique-GRPO
& \ul{55.49} & 37.86 & 77.58 & \ul{36.33} & \ul{85.58} \\
\rowcolor{steelblue!33}
+ \ours
& \textbf{58.49} & \textbf{41.65} & \textbf{80.74} & \textbf{38.33} & \textbf{87.80} \\
\bottomrule
\end{tabular}
}
\end{table}

\paragraph{Evaluation Benchmarks.}
For mathematical reasoning tasks, we evaluate our models on three widely used benchmarks: AIME24, AIME25, AMC23~\citep{numina_math_datasets}.
For instruction following tasks, we evaluate on IFEval~\citep{zhou2023instruction} and IFBench~\citep{pyatkin2025generalizing}.
For code generation, we evaluate on the held-out private tests of the LCBv6 subset of LiveCodeBench~\citep{jain2025livecodebench}.

\paragraph{Implementation Details.}
For mathematical reasoning tasks, we use the Hugging Face \texttt{Math-Verify}\footnote{https://github.com/huggingface/Math-Verify} library to automatically verify the correctness of model answers during both training and evaluation. During evaluation, we sample 8 times for each question and take the average accuracy as the final score. To achieve precise and stable external critique, we employ the indicative critique strategy from \citet{zhang2025critique}, which conditions critiques on ground-truth answers. For instruction-following tasks, we convert code verification functions to natural language, as illustrated in Appendix~\ref{appendix:verify_training_details}. For code generation, we train for 100 steps and report the best Avg@4 accuracy on the held-out private test split throughout training. More experimental details are in Appendix~\ref{appendix:verify_training_details}.

\subsection{Experimental Results}
\label{ssec:verifiable-exp}

\paragraph{\ours achieves the best performance on both mathematical reasoning and instruction following tasks.}

Table~\ref{tab:verifiable_main_results} shows that \ours consistently delivers the strongest results across all verifiable benchmarks for both Qwen-3-4B and Qwen-3-8B. Compared with the GRPO baseline, \ours yields clear gains on mathematical reasoning, improving AIME24 and AIME25 by +6.46 and +2.68 points on Qwen-3-4B, and by +4.44 and +3.63 points on Qwen-3-8B, respectively. The improvements extend beyond math: \ours increases IFBench by +4.34 and +2.68 points and IFEval by +2.06 and +3.04 points over GRPO on the two model sizes. Notably, \ours also outperforms Critique-GRPO across all benchmarks, showing that combining group-level feedback with adaptive guidance provides benefits beyond critique-only refinement.

\begin{figure}[!t]
    \centering
    \definecolor{upurple}{RGB}{155,89,182}
\definecolor{ublue}{RGB}{52,152,219}
\definecolor{ured}{RGB}{231,76,60}
\definecolor{ugreen}{RGB}{46,204,113}

\definecolor{draw1}{RGB}{128, 128, 247}
\definecolor{fill1}{RGB}{179, 179, 250}

\definecolor{red1}{RGB}{246, 214, 226}
\definecolor{yellow1}{RGB}{255, 252, 227}
\definecolor{green1}{RGB}{225, 254, 220}

\begin{tikzpicture}
\scriptsize{
\begin{axis}[
  name=left,
  width=0.8\linewidth,
  height=.22\textwidth,
  scale only axis=true,
  at={(0,0)},
  anchor=west,
  xshift=4.4em,
  ymajorgrids,
  grid style=dashdotted,
  ybar,
  ymin=35, ymax=67,
  ytick={35.00,40,45,50,55,60,65,70},
  yticklabel pos=left,
  ylabel style={font=\scriptsize, yshift=-0.4em,xshift=0.6em},
  yticklabel style={
    /pgf/number format/.cd,
    fixed,
    fixed zerofill,
    precision=2
  },
  ylabel={Average Score (\%)},
  xtick=data,
  symbolic x coords={Fuzzy,Math,IF},
  xticklabels={Non-verifiable, Math, Instruction Following},
  xtick align=inside,
  enlarge x limits=0.25,
  bar width=1.8em,
  nodes near coords,
  nodes near coords align={vertical},
  nodes near coords style={
    font=\tiny,
    scale=0.9,
    /pgf/number format/fixed,
    /pgf/number format/fixed zerofill,
    /pgf/number format/precision=2
  },
  xlabel style={yshift=0.3em,align=center},
  xticklabel style={font=\scriptsize},
  legend style={
    at={(0.30,0.66)},
    anchor=south,
    legend columns=1,
    draw=none,
    fill=none,
    legend image post style={sharp corners},
    cells={anchor=west},
    /tikz/every odd column/.append style={column sep=0.3em},
    /tikz/every even column/.append style={column sep=1em},
  },
  axis on top=false,
]

\addplot[fill=cyan!70, draw=gray, area legend, line width=0.4pt] coordinates {
  (Fuzzy,48.83) (Math,50.07) (IF,63.07)
};
\addlegendentry{\ours}

\addplot[fill=cyan!40, draw=gray, area legend] coordinates {
  (Fuzzy,42.89) (Math,46.75) (IF,61.78)
};
\addlegendentry{w/o intra-group feedback}

\addplot[fill=cyan!10, draw=gray, area legend] coordinates {
  (Fuzzy,39.62) (Math,47.61) (IF,60.81)
};
\addlegendentry{w/o external feedback}

\end{axis}
}
\end{tikzpicture}
    % \vspace{-1.5em}
    \caption{\textbf{Ablation on feedback sources.} We ablate \textit{intra-group attempts} or \textit{external critiques} from the aggregated refinement context. Bars report average performance over the non-verifiable, math reasoning, and instruction following suites and we provide per-benchmark results in Appendix~\ref{appendix:ablation_feedback}.}
    \label{fig:ablation-feedback-type}
\end{figure}
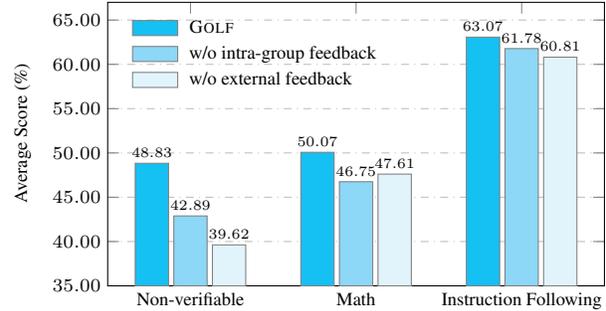
\paragraph{\ours improves both Pass@1 and Pass@k in competition math reasoning benchmarks.}
Figure~\ref{fig:pass_at_k} reports Pass@$k$ (from $k=1$ to $128$) for Qwen-3-8B on two mathematical reasoning benchmarks, AIME25 and AMC23. Across the entire $k$ range, \ours consistently dominates both the base model and the GRPO baseline, indicating more effective search under the same sampling budget. The gains are evident already at small $k$ (higher Pass@1), reflecting improved single-sample quality. More importantly, the advantage persists and becomes more pronounced as $k$ increases, yielding higher Pass@128, which directly signals a richer set of successful solution trajectories. Overall, these trends suggest that group-level feedback-guided refinement improves exploration diversity and helps the policy cover more correct reasoning paths.

% \paragraph{Setup.}
% We adopt LCBv6 subset of LiveCodeBench~\citep{jain2025livecodebench} as our evaluation benchmark, which curates competitive programming problems at varying difficulty levels.
% Following the concurrent work~\citep{hübotter2026reinforcementlearningselfdistillation}, we adopt a public/private unit test setting: we randomly sample 50\% of private tests as public tests, which serve as the verifier during training, while the remaining private tests are held out for evaluation. We use \texttt{Qwen-3-8B} as the base model. We train for 100 steps and report the best Avg@4 accuracy on the private test split throughout training. 
% As baselines, we compare against GRPO following the DAPO setting~\citep{yu2025dapo} with $\varepsilon_{\text{high}}=0.28$, which serves as a strong baseline, and On-Policy Self-Distillation (SDPO;~\citealp{hübotter2026reinforcementlearningselfdistillation}), which utilizes the execution feedback as hindsight information for on-policy distillation.

\definecolor{upink}{HTML}{fcd4d4}
\definecolor{ucyan}{HTML}{e3eeff}
\definecolor{uedgecyan}{HTML}{6d97e0}
\definecolor{uedgepink}{HTML}{cc0000}
\definecolor{upurple}{RGB}{155,89,182}
\definecolor{ublue}{RGB}{52,152,219}
\definecolor{ured}{RGB}{231,76,60}
\definecolor{udark}{RGB}{77,153,77}
\definecolor{ugreen}{RGB}{46,204,113}
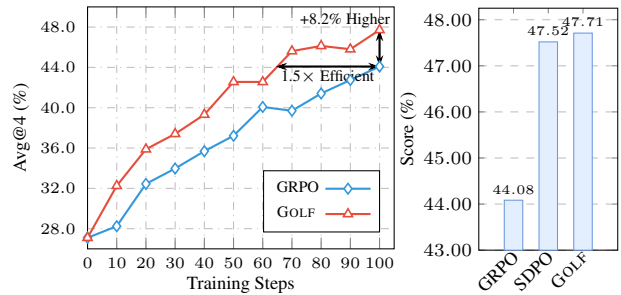
\begin{figure}[!t]
  \centering
  \begin{tikzpicture}

  \begin{axis}[
    name=lineplot,
    at={(-5em,-4.5em)},            
    width=0.33\textwidth, 
    height=0.28\textwidth,
    xlabel=Training Steps,
    ylabel={Avg@4 (\%)},
xtick={0,10,20,30,40,50,60,70,80,90,100},
ytick={28,32,36,40,44,48},
xmin=0,
xmax=105,
ymin=26,ymax=50,
    ymajorgrids=true,
    xmajorgrids=true,
    xticklabel style={font=\tiny},
    yticklabel style={font=\tiny},
    grid style=dashdotted,
    legend cell align=left,
    scaled ticks=false,
    xlabel style={align=center,font=\scriptsize,yshift=0.8em},
    ylabel style={font=\scriptsize,yshift=-1em},
    y tick style={opacity=0},
    legend style={yshift=-6em,xshift=-0.3em,legend cell align=left,legend plot pos=right,font=\tiny},  
    yticklabel style={
      xshift=-0.1em,
      /pgf/number format/.cd,
      fixed,
      fixed zerofill,
      precision=1,
    },
  ]
% GRPO
\addplot [sharp plot,ublue,thick,line width=0.5pt,mark=diamond*,mark size=2pt,thick,mark options={fill=white,draw=ublue,line width=0.6pt}] plot coordinates {
    (0,27.099)(10,28.244)(20,32.443)
    (30,33.969)(40,35.687)(50,37.214)
    (60,40.076)(70,39.695)(80,41.412)
    (90,42.748)(100,44.084)
};

% Ours
\addplot [sharp plot,ured,thick,line width=0.5pt,mark=triangle*,mark size=2pt,thick,mark options={fill=white,draw=ured,line width=0.6pt}] plot coordinates {
    (0,27.099)(10,32.252)
    (20,35.878)(30,37.405)(40,39.313)
    (50,42.557)(60,42.557)(70,45.611)
    (80,46.13)(90,45.802)(100,47.710)
};

\draw[{Stealth[length=1.5mm, width=1mm]}-{Stealth[length=1.5mm, width=1mm]},thick,black] 
    (axis cs:100,44.084) -- (axis cs:100,47.710)
node[midway,above,xshift=-1.3em,yshift=0.4em,font=\tiny,text=black] {+8.2\% Higher};
\draw[{Stealth[length=1.5mm, width=1mm]}-{Stealth[length=1.5mm, width=1mm]},thick,black] (axis cs:100,44.084) -- (axis cs:64.5,44.084)node[midway,above,font=\tiny,xshift=0em,yshift=-1em,text=black] {1.5$\times$ Efficient};
\legend{\tiny{GRPO},\tiny{\ours}},
  \end{axis}

\scriptsize{
\begin{axis}[
    at={(14em,-6.5em)},            
    ymajorgrids,
    grid style=dashed,
    ybar,
    enlarge x limits=0.08,
    xtick align=inside,
    width=0.2\textwidth,
    height=0.21\textheight,
    scaled y ticks = false,
    enlarge y limits={upper,value=0.05},
    bar width=1em,
    enlarge x limits=0.5,
    ylabel=Score (\%),
    xtick={2,3,4},
    xticklabels={GRPO,SDPO,\ours},
    ymin=43, ymax=48,
    ytick={43.00,44.00,45.00,46.00,47.00,48.00,49.00},
    yticklabel pos=left,
    xlabel style={yshift=0.3em,align=center},
    xticklabel style={                
    rotate=45,
    anchor=north east,
    font=\scriptsize,
    yshift=0.5em,
    xshift=0.3em
},
    ylabel style={font=\scriptsize,yshift=-1.2em},
    yticklabel style={
    font=\scriptsize,
    /pgf/number format/.cd,
    fixed,
    fixed zerofill,
    precision=2
},
    axis on top=false,
    major tick length=2pt,
]
\addplot+[
    ybar,
    fill=ucyan,
    draw=uedgecyan,
    area legend,
    point meta=y,
    nodes near coords={%
        \pgfmathprintnumber[fixed,precision=2,zerofill]{\pgfplotspointmeta}%
    },
    nodes near coords style={
        font=\fontsize{4}{5}\selectfont,
        anchor=south,
        text=black,
    }
] coordinates { 
    (2,44.084)
    (3,47.519)
    (4,47.71)
};
\end{axis}}

  \end{tikzpicture}
  \caption{Evaluation on LCBv6 with \texttt{Qwen-3-8B}. \textbf{Left:} Avg@4 accuracy curve over training steps. \textbf{Right:} Final performance comparison with SDPO~\citep{hübotter2026reinforcementlearningselfdistillation} under the same environment feedback setting.}
  \label{fig:code-eval}
\end{figure}

\paragraph{\ours extends gains to code generation with rich 
environment feedback.}
Beyond binary correctness signals, coding environments provide richer natural language feedback such as runtime errors and failed unit tests, making them a particularly informative testbed for \ours's group-level feedback aggregation. As shown in Figure~\ref{fig:code-eval} (left), \ours achieves an Avg@4 of \textbf{47.71} on LCBv6, outperforming the GRPO baseline by \textbf{+3.63} points while achieving 1.5$\times$ sample efficiency. In addition, \ours also slightly outperforms SDPO (47.71 vs.\ 47.51),  despite the two methods operating from fundamentally different angles: SDPO leverages execution feedback together with successful attempts to provide dense rewards for the policy's rollouts, whereas \ours aggregates execution feedback with diverse failure patterns to construct targeted refinement guidance. As the two approaches exploit complementary signals, namely past successes versus diverse failures, their combination presents a promising direction for future work. 
\section{Ablation Study and Analysis}
We perform ablation studies on the key design choices of \ours: (1) group-level feedback; (2) adaptive guidance; and (3) joint optimization for self-refinement. We also provide ablation studies on training efficiency in Appendix~\ref{appendix:efficiency}.

\subsection{Effect of Group-level Feedback}
Our framework relies on group-level NL feedback that combines \emph{external critiques} with \emph{intra-group attempts}.
We ablate each source: \textbf{w/o external feedback} builds refinement prompts from failure attempts only, while \textbf{w/o intra-group attempts} refines a single sampled response using its critique.
Figure~\ref{fig:ablation-feedback-type} shows that removing either component consistently harms performance across all task types.
On non-verifiable tasks, removing intra-group attempts and critiques leads to a \textbf{12.2\%} and \textbf{18.9\%} drop, respectively; on mathematical reasoning, the drops are \textbf{6.6\%} and \textbf{4.9\%}; and on instruction following, \textbf{2.0\%} and \textbf{3.6\%}.
These results highlight the complementarity of the two feedback sources: critiques provide targeted revision signals, whereas alternative attempts supply reusable partial ideas and diverse failure patterns, and their combination yields higher-quality refinements.
We further provide a case study in Appendix~\ref{appendix:case_study} to illustrate the effectiveness of group-level feedback.

\subsection{Ablation on Adaptive Guidance}
A key design choice in \ours is how to leverage high-quality refinements derived from NL feedback to guide exploration.
This design comprises two aspects: (i) \emph{how} to learn from off-policy refinements, and (ii) \emph{when} to inject them.

\paragraph{How to learn from off-policy refinements.}
\definecolor{upink}{HTML}{fcd4d4}
\definecolor{ucyan}{HTML}{e3eeff}
\definecolor{uedgecyan}{HTML}{6d97e0}
\definecolor{uedgepink}{HTML}{cc0000}
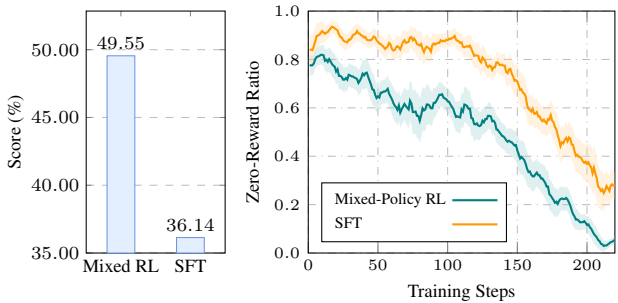
\begin{figure}[!t]
  \centering
  \begin{tikzpicture}

\scriptsize{

\begin{axis}[
    at={(-5em,-4.5em)},
    ymajorgrids,
    grid style=dashed,
    ybar,
    enlarge x limits=0.08,
    xtick align=inside,
    width=0.2\textwidth,
    height=0.21\textheight,
    scaled y ticks = false,
    enlarge y limits={upper,value=0.05},
    bar width=1.5em,
    enlarge x limits=0.5,
    ylabel=Score (\%),
    xtick={2,4},
    xticklabels={Mixed RL,SFT},
    ymin=35, ymax=52,
    ytick={35.00,40.00,45.00,50.00},
    yticklabel pos=left,
    xlabel style={yshift=0.3em,align=center},
    ylabel style={font=\scriptsize,yshift=-1.2em},
    yticklabel style={
    font=\scriptsize,
    /pgf/number format/.cd,
    fixed,
    fixed zerofill,
    precision=2
},
    axis on top=false,
    major tick length=2pt,  
]
\addplot+[
    ybar,
    fill=ucyan,
    draw=uedgecyan,
    area legend,
    point meta=y,
    nodes near coords={%
        \pgfmathprintnumber[fixed,precision=2,zerofill]{\pgfplotspointmeta}%
    },
    nodes near coords style={
        font=\scriptsize,
        anchor=south,
        text=black,
    }
] coordinates { 
    (2,49.55)
    (4,36.14)
};
\end{axis}}

  \begin{axis}[
    name=lineplot,
    at={(7em,-4.5em)},
    width=0.33\textwidth, 
    height=0.28\textwidth,
    xlabel=Training Steps,
    ylabel=Zero-Reward Ratio,
    xmin=0, xmax=220,
    ymin=0, ymax=1.0,
    xtick={0,50,100,150,200},
    ytick={0,0.2,0.4,0.6,0.8,1.0},
    ymajorgrids=true,
    xmajorgrids=true,
    xticklabel style={font=\tiny},  
    yticklabel style={font=\tiny},  
    grid style=dashdotted,
    legend cell align=left,
    scaled ticks=false,
    xlabel style={align=center,font=\scriptsize,yshift=0.8em},
    ylabel style={font=\scriptsize,yshift=-2em},
    y tick style={opacity=0},
    legend style={yshift=-9em,xshift=-5.5em,legend cell align=left,legend plot pos=right,font=\tiny},
    yticklabel style={
      xshift=-0.1em,
      /pgf/number format/.cd,
      fixed,
      fixed zerofill,
      precision=1,
    },
  ]
    % Mixed-Policy RL: 阴影区域
    \addplot [name path=upper_bound1, draw=none, forget plot] 
      table [x=step, y=upper_bound, col sep=comma] {figures/zero_reward_trend_mixed-policy_rl.csv};
    \addplot [name path=lower_bound1, draw=none, forget plot] 
      table [x=step, y=lower_bound, col sep=comma] {figures/zero_reward_trend_mixed-policy_rl.csv};
    \addplot [teal!40, opacity=0.3, forget plot] fill between[of=upper_bound1 and lower_bound1];
    
    % Mixed-Policy RL: 主曲线
    \addplot [sharp plot, teal, thick, line width=0.8pt] 
      table [x=step, y=smoothed_ratio, col sep=comma] {figures/zero_reward_trend_mixed-policy_rl.csv};
    \addlegendentry{Mixed-Policy RL}

    % SFT: 阴影区域
    \addplot [name path=upper_bound2, draw=none, forget plot] 
      table [x=step, y=upper_bound, col sep=comma] {figures/zero_reward_trend_sft.csv};
    \addplot [name path=lower_bound2, draw=none, forget plot] 
      table [x=step, y=lower_bound, col sep=comma] {figures/zero_reward_trend_sft.csv};
    \addplot [orange!40, opacity=0.3, forget plot] fill between[of=upper_bound2 and lower_bound2];
    
    % SFT: 主曲线
    \addplot [sharp plot, orange, thick, line width=0.8pt] 
      table [x=step, y=smoothed_ratio, col sep=comma] {figures/zero_reward_trend_sft.csv};
    \addlegendentry{SFT}
    
  \end{axis}
  \end{tikzpicture}
  \caption{Left: Average performance comparison of different off-policy sample learning strategies on non-verifiable benchmarks; full results are provided in Appendix~\ref{tab:ablation_sft}. Right: Zero-reward ratio during training.}
  \label{fig:mixed-policy-rl}
\end{figure}
To assess the effectiveness of mixed-policy optimization, we compare \textit{mixed-policy RL}---which injects high-quality refinements into the on-policy group $\mathcal{G}_{\text{gen}}(x)$ and optimizes it with Eq.~\ref{eq:mixed_obj}---against \textit{Supervised Fine-Tuning (SFT)}, which trains the policy to imitate the same refinement samples with a supervised learning objective.
Experiments on non-verifiable benchmarks in Figure~\ref{fig:mixed-policy-rl} (left) show that \textit{mixed-policy RL} consistently outperforms \textit{SFT}, with an average improvement of 37.10\%.
Moreover, we track the fraction of all-zero-reward groups during training. As shown in Figure~\ref{fig:mixed-policy-rl} (right), when applying mixed-policy optimization, \ours markedly reduces the frequency of all-zero groups by 31.97\%, thereby yielding usable policy gradients. In contrast, SFT improves imitation of refinement outputs but does not reliably propagate these improvements to on-policy exploration, leading to slower recovery from low-reward regimes.

\definecolor{mygrey}{HTML}{E6E6E6}

\begin{table}[t]
\centering
\caption{RefineBench performance under two settings: \emph{self-refinement}, where the model refines its initial response, and \emph{guided-refinement}, where the refinement is conditioned on the checklists.}
\label{tab:refine_bench}
\resizebox{\columnwidth}{!}{%
\begin{tabular}{lcccc}
\toprule
\textbf{Model} & \textbf{Biology} & \textbf{Chemistry} & \textbf{Law} & \textbf{Average} \\
\midrule
\rowcolor{mygrey}
\multicolumn{5}{l}{\textbf{Self-Refinement}} \\
% \midrule
\textbf{\textit{Llama-3.1-8B-Instruct}} & 9.09 & 8.33 & 23.94 & 13.79 \\
+ GRPO & 22.73 & 27.77 & 22.24 & 24.25 \\
+ \ours & \textbf{27.27} & \textbf{33.33} & \textbf{24.64} & \textbf{28.41} \\

\midrule
\rowcolor{mygrey}
\multicolumn{5}{l}{\textbf{Guided-Refinement}} \\
% \midrule
\textbf{\textit{Llama-3.1-8B-Instruct}} & 18.18 & 27.78 & 49.30 & 31.75 \\
+ GRPO & 27.78 & 55.55 & 45.07 & 42.80 \\
+ \ours & \textbf{54.55} & \textbf{66.67} & \textbf{50.00} & \textbf{57.07} \\

\bottomrule
\end{tabular}
}
\end{table}

\paragraph{When to inject refinements.}
We further ablate the injection schedule by comparing our \textit{adaptive injection} strategy, which is triggered only when the generation group $\mathcal{G}_{\text{gen}}$ is in a low-reward regime, with an \textit{always injection} variant that injects the highest-reward refinements for each rollout group.
As shown in Table~\ref{tab:ablation_injection}, adaptive injection yields the best results, outperforming the \textit{always injection} strategy by 27.37\%.
Intuitively, it allocates off-policy guidance to the prompts where GRPO is most likely to suffer from collapsed group-normalized advantages (e.g., all-zero groups), thereby converting previously uninformative groups into ones with usable gradients.
In contrast, always injecting may dilute the benefit of scaffolding, leading to weaker overall gains.

\subsection{Effect of Joint Optimization for Self-Refinement}
We further analyze the impact of jointly training for self-refinement on the model's refinement capabilities.
We evaluate \ours on RefineBench~\citep{lee2025refinebenchevaluatingrefinementcapability}, which measures refinement capability via checklists under two settings: \emph{self-refinement} and \emph{guided refinement}.
Table~\ref{tab:refine_bench} shows the \texttt{Pass} ratio, where we assign a score of 1 only if all checklist items are correct; otherwise, we assign 0.
As shown, \ours consistently improves refinement performance over GRPO in both settings.
On average, \ours increases the pass rate from 24.25 to 28.41 in self-refinement and from 42.80 to 57.07 in guided refinement.
Notably, optimizing only for problem solving with GRPO can be misaligned with refinement: GRPO does not reliably improve and can even degrade refinement performance on some domains (\eg, Law in both settings), whereas \ours maintains consistent gains.
These results indicate that jointly optimizing generation and refinement within a unified RL process is important for developing robust test-time self-refinement, and for better utilizing explicit NL feedback when available.
\subsection{Entropy Analysis}
\begin{figure}[!t]
  \centering
  \begin{tikzpicture}
  \scriptsize

  \begin{axis}[
    width=0.5\textwidth, height=0.28\textwidth,
    xlabel=Training Steps,
    ylabel=Entropy,
    xmin=0, xmax=224,
    ymin=0, ymax=940,
    xtick={0,50,100,150,200},
    ytick={0,100.0,200,300,400,500,600,700,800,900},
    ymajorgrids=true,
    xmajorgrids=true,
    grid style=dashdotted,
    legend cell align=left,
    scaled ticks=false,
    xlabel style={align=center,font=\footnotesize,yshift=1em},
    ylabel style={font=\footnotesize,yshift=-1em},
    y tick style={opacity=0},
    legend style={yshift=-0.5em,xshift=0em,legend cell align=left,legend plot pos=right},
    yticklabel style={
    xshift=-0.2em,  
    /pgf/number format/.cd,
    fixed,
    fixed zerofill,
    precision=0,
  },
  ]

    % Ours: 阴影区域
    \addplot [name path=Ours_upper, draw=none, forget plot] 
      table [x=Step, y=Ours_upper, col sep=comma] {figures/entropy2-smooth-shixian.csv};
    \addplot [name path=Ours_lower, draw=none, forget plot] 
      table [x=Step, y=Ours_lower, col sep=comma] {figures/entropy2-smooth-shixian.csv};
    \addplot [red!40, opacity=0.3, forget plot] fill between[of=Ours_upper and Ours_lower];
    
    % Ours: 主曲线
    \addplot [sharp plot, red, thick, line width=0.8pt] 
      table [x=Step, y=Ours, col sep=comma] {figures/entropy2-smooth-shixian.csv};
    \addlegendentry{\scriptsize{\ours}}

    % Baseline: 阴影区域
    \addplot [name path=baseline_upper, draw=none, forget plot] 
      table [x=Step, y=baseline_upper, col sep=comma] {figures/entropy2-smooth-shixian.csv};
    \addplot [name path=baseline_lower, draw=none, forget plot] 
      table [x=Step, y=baseline_lower, col sep=comma] {figures/entropy2-smooth-shixian.csv};
    \addplot [blue!40, opacity=0.3, forget plot] fill between[of=baseline_upper and baseline_lower];
    
    % Baseline: 主曲线
    \addplot [sharp plot, blue, thick, line width=0.8pt] 
      table [x=Step, y=baseline, col sep=comma] {figures/entropy2-smooth-shixian.csv};
    \addlegendentry{\scriptsize{Pairwise-GRPO}}
    
  \end{axis}

  \end{tikzpicture}
  \caption{Policy entropy (seq-mean-token-sum-norm) of \ours and the Pairwise-GRPO baseline over training steps.}
  \label{fig:entropy}
\end{figure}
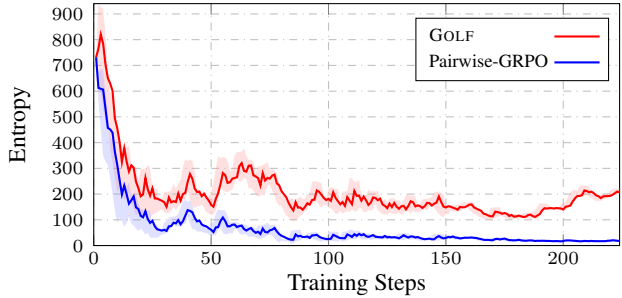
Entropy has long been regarded as a proxy for exploration in policy optimization. To assess the impact of \ours on exploratory behavior, we track the policy entropy of \texttt{Llama-3.1-8B-Instruct} throughout RL training on non-verifiable tasks. Figure~\ref{fig:entropy} shows that \ours maintains consistently higher entropy than the Pairwise-GRPO baseline. While the baseline exhibits a rapid entropy collapse, \ours stabilizes at a substantially higher level and displays recurrent entropy surges over training, suggesting sustained exploration rather than premature mode collapse. This trend aligns with our design: aggregating intra-group attempts exposes complementary partial ideas and diverse failure modes, helping the policy escape local optima and continue exploring diverse trajectories.
\section{Conclusion}
We presented a natural language feedback guided RL framework that improves RL exploration by turning rich textual feedback into actionable training signals. Our core idea is to use refinement and critique as guidance to densify learning signals when scalar rewards are sparse, while aggregating feedback at the group level to alleviate information bottlenecks and broaden the space of explored behaviors. Across non-verifiable and verifiable tasks, our method achieves substantially faster convergence and higher trajectory diversity, demonstrating its effectiveness in improving RL exploration. Ablations further confirm the complementary roles of intra feedback and critique in driving these benefits. Overall, our results suggest that natural language guidance provides a practical and scalable path to more efficient and diverse exploration in language model reinforcement learning.

\section*{Impact Statement}

This work aims to improve reinforcement learning for large language models by exploiting natural language feedback to guide exploration and refinement, ultimately improving training efficiency and final performance. A potential benefit is reduced trial-and-error during RL training, which may lower computational cost and improve reliability in interactive settings where iterative correction is common.
However, stronger refinement and exploration may also amplify risks: models could become more effective at generating persuasive or strategically optimized content, and LLM-based judges or critiques may introduce bias that is then reinforced during training. We encourage careful evaluation, bias auditing of judge feedback, and responsible deployment practices when applying these methods in real-world systems.

\nocite{langley00}

\bibliography{example_paper}
\bibliographystyle{icml2026}
\newpage
\appendix
\onecolumn
\section{Preliminary Study: The Value of Diverse Feedback}
\label{appendix:preliminary}

A core motivation of \ours is that NL feedback encompasses more than explicit critiques. In practice, alternative attempts within a rollout group can also serve as implicit guidance by exposing diverse failure modes and partial solution strategies. To validate this intuition, we conduct a preliminary study addressing two hypotheses: \textbf{(H1)} external critiques improve test-time self-refinement; \textbf{(H2)} combining intra-group attempts with critiques yields additional gains by helping the model escape local minima and explore more diverse repair directions.

\subsection{Experimental Setup}

We focus on mathematical reasoning and construct a challenging subset 
from OpenR1~\citep{openr1math}. Specifically, we filter 500 problems 
on which \texttt{Qwen-3-8B} achieves zero pass@4 
accuracy—that is, the model fails to produce any correct solution 
across four independent samples per problem. For each problem, we 
perform refinement under different feedback conditions and evaluate 
using two metrics: \textbf{pass@4}, the fraction of problems solved 
by at least one refined sample (out of four), and \textbf{Acc}, the 
overall accuracy across all 2,000 refined outputs (500 problems 
$\times$ 4 samples).

\subsection{Feedback Conditions}

We compare four conditions with increasing information richness:

\noindent\textbf{Simple.} The model receives only a binary signal indicating the current solution is incorrect, with no diagnostic information.

\noindent\textbf{Intra-Feedback.} The model is provided with other failed attempts from the same rollout group, enabling it to reuse partial reasoning steps and avoid repeated mistakes, but without explicit error analysis.

\noindent\textbf{External-Feedback.} The model receives a critique for the current response, generated by \texttt{Qwen3-235B-A22B-Instruct-2507}, which pinpoints concrete errors and suggests revisions.

\noindent\textbf{Mixed Feedback.} We combine intra-group attempts with external critiques, providing both targeted diagnostics and alternative reasoning traces.

\subsection{Results and Analysis}

\begin{table}[htbp]
\centering
\caption{Test-time refinement results under different feedback conditions.}
\label{tab:preliminary}
\begin{tabular}{lcc}
\toprule
\textbf{Feedback Type} & \textbf{pass@4 (\%)} & \textbf{Acc (\%)} \\
\midrule
Simple       & 0.00  & 0.00  \\
Intra        & 18.80 & 6.65  \\
External     & 27.60 & 17.00 \\
Mixed        & \textbf{30.40} & \textbf{17.55} \\
\bottomrule
\end{tabular}
\end{table}

Table~\ref{tab:preliminary} summarizes the results. External feedback yields substantial improvements, lifting pass@4 from 0\% to 27.60\% and Acc from 0\% to 17.00\%, confirming \textbf{(H1)}: explicit critiques provide strong corrective signals. Intra-group feedback alone also improves pass@4 to 18.80\%, though the gain in Acc is more modest (6.65\%). This asymmetry suggests that alternative attempts primarily help broaden the search space rather than directly fixing specific errors—models may generate more diverse solutions but still struggle to identify which partial ideas are correct.

Crucially, mixed feedback achieves the best performance (30.40\% pass@4, 17.55\% Acc), outperforming external feedback alone by +2.80\% in pass@4 and +0.55\% in Acc. This supports \textbf{(H2)}: the two sources are complementary. External critiques provide targeted revision directions, while intra-group attempts supply diverse reasoning fragments that help escape unproductive trajectories. Notably, the improvement in Acc (not just pass@k) indicates that combining both sources produces more consistently correct solutions, not merely more lucky guesses.

These findings motivate our design choice to aggregate both external critiques and intra-group attempts at the group level, jointly leveraging their complementary strengths to generate higher-quality training data for policy learning.

\section{Prompts for Self-Refinement}
\label{appendix:refine_prompts}
In this section, we present the prompts used for model self-refinement during RL training. Depending on the task and NL feedback, we employ different prompts.
\subsection{Non-verifiable Tasks}
\begin{tcolorbox}[promptbox,
  title=Self-Refinement Prompt for WildChat,breakable]
  \small

\begin{verbatim}
Given the following inputs:

**Problem**: {original_prompt}

**Candidate Responses with Feedback**:

--- Candidate Response 1 (Score: {score_1}) ---
Response:
{response_1}

Feedback:
{critique_1}

--- Candidate Response 2 (Score: {score_2}) ---
Response:
{response_2}

Feedback:
{critique_2}

--- Candidate Response 3 (Score: {score_3}) ---
Response:
{response_3}

Feedback:
{critique_3}

... (additional candidates if provided)

Please synthesize an improved response by:

- Learning from the mistakes identified in the critiques - avoid repeating the same errors.
- Incorporating the strengths and good aspects mentioned in the critiques.
- Synthesizing the best parts from all candidates while addressing their individual weaknesses.
- Fully satisfying the user instruction and meeting all requirements.

CRITICAL OUTPUT REQUIREMENTS:
- Provide ONLY the synthesized response itself, nothing more.
- DO NOT start with any meta phrases like "Improved Response:", "Here is the synthesized response:", or similar introductory text.
- DO NOT end with any meta commentary, notes, or explanations such as "Note: This response meets all requirements...", "This addresses the user's needs...", or any other additional remarks.
- Your output should be the pure, direct response to the user's original instruction - as if you were directly answering them without any wrapper text or self-commentary.
\end{verbatim}

\end{tcolorbox}
\subsection{Verifiable Tasks}

\begin{tcolorbox}[promptbox,
  title=Self-Refinement Prompt for Instruction Following,breakable]
  \small

\begin{verbatim}
Given the following inputs:

**User Instruction**: {original_prompt}

**Candidate Responses with Feedback**:

--- Candidate Response 1 (Score: {score_1}) ---
Response:
{response_1}

Feedback:
{critique_1}

--- Candidate Response 2 (Score: {score_2}) ---
Response:
{response_2}

Feedback:
{critique_2}

--- Candidate Response 3 (Score: {score_3}) ---
Response:
{response_3}

Feedback:
{critique_3}

... (additional candidates if provided)

Please synthesize an improved response by:

- Learning from the mistakes identified in the critiques - avoid repeating the same errors.
- Incorporating the strengths and good aspects mentioned in the critiques.
- Synthesizing the best parts from all candidates while addressing their individual weaknesses.
- Fully satisfying the user instruction and meeting all requirements.

CRITICAL OUTPUT REQUIREMENTS:
- Provide ONLY the synthesized response itself, nothing more.
- DO NOT start with any meta phrases like "Improved Response:", "Here is the synthesized response:", or similar introductory text.
- DO NOT end with any meta commentary, notes, or explanations such as "Note: This response meets all requirements...", "This addresses the user's needs...", or any other additional remarks.
- Your output should be the pure, direct response to the user's original instruction - as if you were directly answering them without any wrapper text or self-commentary.
\end{verbatim}

\end{tcolorbox}

\begin{tcolorbox}[promptbox,
  title=Refinement Prompt for Mathematical Reasoning,breakable]
  \small

\begin{verbatim}
Given the following inputs:

**Problem**: {original_prompt}

**Solution Attempts with Feedback**:

--- Solution Attempt 1 (Score: {score_1}) ---
{solution_1}
Feedback: The answer is incorrect, the ground_truth is {ground_truth}

--- Solution Attempt 2 (Score: {score_2}) ---
{solution_2}
Feedback: The answer is incorrect, the ground_truth is {ground_truth}

--- Solution Attempt 3 (Score: {score_3}) ---
{solution_3}
Feedback: The answer is incorrect, the ground_truth is {ground_truth}

... (additional attempts if provided)

Please synthesize an improved solution by:

- Carefully analyze each attempt and use the feedback to locate where the reasoning goes wrong.
- Keep any valid steps and calculations, but fix incorrect steps with your own correct reasoning.
- Build a complete, coherent, self-contained solution with step-by-step derivations and necessary calculations.

CRITICAL REQUIREMENTS:
- You MUST derive the solution through genuine mathematical reasoning.
- Do NOT work backwards from the ground truth or force the final answer.
- Every step must follow logically from the previous one and use valid mathematical operations.

OUTPUT FORMAT:
- Start solving immediately (no preface or meta commentary).
- Do NOT mention or allude to attempts, candidates, or feedback.
- End with the final answer formatted as \boxed{}.
- Output ONLY the solution itself. Do not add any notes after the boxed answer.
\end{verbatim}

\end{tcolorbox}

\section{Training Details for Non-verifiable Tasks}
\label{appendix:fuzzy_training_details}
This section describes the training details for non-verifiable tasks. We provide the training hyperparameters in Table~\ref{tab:hparams_non_verifiable}.

\section{Training Details for Verifiable Tasks}
\label{appendix:verify_training_details}
This section describes the training details for verifiable tasks. Table~\ref{tab:hparams_math} lists the training hyperparameters for mathematical reasoning tasks, and Table~\ref{tab:hparams_if} lists the training hyperparameters for instruction-following tasks.
For mathematical reasoning and verifiable instruction-following tasks, the sources of feedback used during training differ from those used for non-verifiable tasks, where a generative reward model is employed.

For mathematical reasoning tasks, providing accurate critiques at the level of intermediate reasoning steps is challenging. Following prior work, we therefore directly use the ground-truth answer as external critique. In addition to indicating whether the model’s prediction is correct, the correct answer is explicitly provided, enabling the model to reflect on and revise incorrect reasoning trajectories.

For instruction-following tasks, the verification process is implemented through code-based constraint checking. We convert the execution results of these verification functions into natural language feedback. Below, we provide an illustrative example: the \texttt{check\_following} function performs automated constraint verification, and based on its pass or fail outcome, the corresponding \texttt{get\_critique} function returns explicit natural language feedback to guide the model during training.
\begin{tcolorbox}[promptbox,
  title=Example: Code-Based Verification and Natural Language Critique (Unique Word Count),breakable]
\small
\begin{verbatim}
def check_following(self, value):
    """Checks if the response contains the expected number of unique words."""
    words = value.lower().split()
    unique_words = set()
    for word in words:
        unique_words.add(word.strip(''.join(string.punctuation) + ' '))
    # Convert to set to get unique words
    return len(unique_words) >= self._num_unique_words

def get_critique(self, value, passed):
    """Generate natural language feedback for unique word count constraint."""
    words = value.lower().split()
    unique_words = set()
    for word in words:
        unique_words.add(word.strip(''.join(string.punctuation) + ' '))
    actual_unique = len(unique_words)
    if passed:
        return f"Constraint satisfied: The response contains {actual_unique} unique words (at least {self._num_unique_words} required)."
    else:
        return f"Constraint not satisfied: The response contains {actual_unique} unique words, but at least {self._num_unique_words} unique words are required."
\end{verbatim}
\end{tcolorbox}

\section{Benchmarks}
\label{appendix:benchmarks}
% \paragraph{Non-verifiable Benchmarks.}
We provide detailed descriptions and statistics of the benchmarks in our non-verifiable task experiments (\S\ref{sec:fuzzy}).
\begin{itemize}
    \item \textbf{AlpacaEval-v2} contains 805 user prompts paired with reference responses. It reports a head to head win rate computed by a generative judge. Following the recommended protocol, we use the length controlled win rate, while replacing the default judge with GPT 4o.
    \item \textbf{WildBench} measures open domain conversational ability using 1{,}024 user prompts, including a subset of multi turn interactions. It is scored with instance level rubrics that are manually checked, which helps reduce judge shortcutting. For each prompt, a candidate response is compared against a GPT 4 reference and assigned a discrete score in $\{-100,-50,0,50,100\}$. We report the mean score over all instances.
    \item \textbf{ArenaHard-v2} consists of 500 challenging real world user queries. We adopt the evaluation setting that uses a GPT 4.1 judge with style control to mitigate potential bias.
    \item \textbf{CreativeWriting-v3} evaluates long form writing under explicit constraints using 96 story chapters. We compute an absolute score between 0 and 100 using GPT 4.1 as the judge.
    \item \textbf{IFEval}: Comprises 25 verifiable instruction types (length, keyword, format, language) with over 500 prompts. Verification is automated with Python functions, supporting strict and loose accuracy metrics.
    \item \textbf{IFBench}: Introduces 58 diverse constraints, curated for their novelty and coverage; all have corresponding verification code. It explicitly targets generalization by constructing evaluation prompts and constraints out-of-domain relative to training data, revealing severe overfitting in prior models.
\end{itemize}
\begin{table}[h]
\centering
\caption{Key hyperparameters used for RL training for non-verifiable tasks in the verl~\citep{sheng2025hybridflow} framework. }
\label{tab:hparams_non_verifiable}
\begin{tabular}{@{}lll@{}}
\toprule
\textbf{Category} & \textbf{Hyperparameter} & \textbf{Value} \\
\midrule
\multirow{4}{*}{Data} 
    & Train file & WildChat-IF \\
    & Max prompt length & 8192 \\
    & Max response length & 4096 \\
    & Filter overlong prompts & True \\
\midrule
\multirow{6}{*}{Actor Model} 
    & Base model 1 & \texttt{Llama-3.1-8B-Instruct} \\
    & Base model 2 & \texttt{Qwen-3-8B} \\
    & LR & $1 \times 10^{-6}$ \\
    & KL loss coefficient $\beta$ & 0.00 \\
    & Use dynamic batch size & True \\
\midrule
\multirow{4}{*}{Rollout} 
    & Rollout engine & vllm \\
    & GPU mem utilization & 0.8 \\
    & Train rollout n & 8\\
    & Temperature & 1.0 \\
\midrule
\multirow{1}{*}{Reward Model} 
    & RM model & \texttt{Qwen3-235B-Instruct-A22B} \\
\midrule
\multirow{6}{*}{Trainer} 
    & PPO Mini Batch size & 32 \\
    & PPO Train Batch size & 64 \\
    & Critic Warmup & 0 \\
    & GPUs per node & 8 \\
    & Nodes & 4\\
    & Total epochs & 2 \\
\bottomrule
\end{tabular}
\end{table}
\begin{table}[h]
\centering
\caption{Key hyperparameters used for RL training for the mathematical reasoning task used in the verl~\citep{sheng2025hybridflow} framework. }
\label{tab:hparams_math}
\begin{tabular}{@{}lll@{}}
\toprule
\textbf{Category} & \textbf{Hyperparameter} & \textbf{Value} \\
\midrule
\multirow{4}{*}{Data} 
    & Train file & OpenR1-filtered \\
    & Max prompt length & 8192 \\
    & Max response length & 6144 \\
    & Filter overlong prompts & True \\
\midrule
\multirow{6}{*}{Actor Model} 
    & Base model 1 & \texttt{Qwen-3-4B (non-thinking)} \\
    & Base model 2 & \texttt{Qwen-3-8B (non-thinking)} \\
    & LR & $1 \times 10^{-6}$ \\
    & KL loss coefficient $\beta$ & 0.00 \\
    & Use dynamic batch size & True \\
\midrule
\multirow{4}{*}{Rollout} 
    & Rollout engine & vllm \\
    & GPU mem utilization & 0.8 \\
    & Train rollout n & 8\\
    & Temperature & 1.0 \\
\midrule
\multirow{1}{*}{Reward Model} 
    & RM model & \texttt{Math-Verify} \\
\midrule
\multirow{6}{*}{Trainer} 
    & PPO Mini Batch size & 128 \\
    & PPO Train Batch size & 64 \\
    & Critic Warmup & 0 \\
    & GPUs per node & 8 \\
    & Nodes & 4\\
    & Total epochs & 10 \\
\bottomrule
\end{tabular}
\end{table}
\begin{table}[H]
\centering
\caption{Key hyperparameters used for RL training for the instruction following task used in the verl~\citep{sheng2025hybridflow} framework. }
\label{tab:hparams_if}
\begin{tabular}{@{}lll@{}}
\toprule
\textbf{Category} & \textbf{Hyperparameter} & \textbf{Value} \\
\midrule
\multirow{4}{*}{Data} 
    & Train file & IFTrain-filtered \\
    & Max prompt length & 8192 \\
    & Max response length & 4096 \\
    & Filter overlong prompts & True \\
\midrule
\multirow{6}{*}{Actor Model} 
    & Base model 1 & \texttt{Qwen-3-4B (non-thinking)} \\
    & Base model 2 & \texttt{Qwen-3-8B (non-thinking)} \\
    & LR & $1 \times 10^{-6}$ \\
    & KL loss coefficient $\beta$ & 0.00 \\
    & Use dynamic batch size & True \\
\midrule
\multirow{4}{*}{Rollout} 
    & Rollout engine & vllm \\
    & GPU mem utilization & 0.8 \\
    & Train rollout n & 8\\
    & Temperature & 1.0 \\
\midrule
\multirow{1}{*}{Reward Model} 
    & RM model & \texttt{Verification functions in Python} \\
\midrule
\multirow{6}{*}{Trainer} 
    & PPO Mini Batch size & 256 \\
    & PPO Train Batch size & 256 \\
    & Critic Warmup & 0 \\
    & GPUs per node & 8 \\
    & Nodes & 4\\
    & Total epochs & 15 \\
\bottomrule
\end{tabular}
\end{table}
\section{LLM Judge Prompts}
\label{appendix:prompts}
In this section, we present the Generative Reward Model scoring prompts used by different baselines on non-verifiable tasks, including pairwise comparison against a high-quality reference (\S\ref{appendix:pairwise_prompt}), rubric-based scoring (\S\ref{appendix:rubric_prompts}), and Likert scoring (\S\ref{appendix:likert_prompt}).
\subsection{Pairwise Scoring Prompt}
\label{appendix:pairwise_prompt}
\begin{tcolorbox}[promptbox,
  title=Pairwise GRM Scoring Prompt,breakable]
  \small
\begin{verbatim}
You are given a user question and two responses.
- Response A: a model-generated answer that may need improvement.
- Response B: a high-quality reference answer (for evaluation only).

Your task is to act as an impartial judge and decide which response is better for the user.

First, think step by step and put your analysis in <reasoning> and </reasoning> tags. In your reasoning:
- Identify the key requirements of the user question (instruction following, relevance, completeness, factuality/safety, style/format).
- Compare Response A and Response B on these requirements.
- Response A may be better than Response B if it follows the user more closely, is safer, or better matches the requested style.
- Avoid position bias and length bias.

Then, provide an actionable critique in <critique> and </critique> tags **for the model-generated answer (Response A) only**. This critique will be shown to another model that only sees that answer. Therefore:
- Write the critique as if there is ONLY ONE answer.
- Do NOT mention or hint that there was another response, a reference answer, "Assistant A/B", "the other answer", "the second answer", or "the reference".
- Point out the current answer’s strengths.
- Point out missing/incorrect/unsafe/irrelevant parts.
- Give concrete suggestions on what to add/remove/fix.
- Do NOT copy or paraphrase the content of Response B.

Finally, output your verdict in <answer> and </answer> tags:
- <answer> [[A]] </answer> if the model-generated answer (Response A) is better
- <answer> [[B]] </answer> if the reference answer (Response B) is better

Format your output EXACTLY like this:
<reasoning> your step-by-step comparison of A vs B </reasoning>
<critique>
your self-contained, neutral feedback for improving the model-generated answer
</critique>
<answer> [[A]] or [[B]] </answer>

Below are the user’s question and the two responses:

[User Question]
{instruction}

[The Start of Response A]
{response_a}
[The End of Response A]

[The Start of Response B]
{response_b}
[The End of Response B]
\end{verbatim}
\end{tcolorbox}

\subsection{Rubrics-as-Reward Scoring Prompt}
\label{appendix:rubric_prompts}
We present the prompts used for rubric generation, as well as the prompts for rubric-based scoring.
\begin{tcolorbox}[promptbox,
  title=Rubric Generation Prompt,breakable]
  \small

You are an expert rubric writer. Your job is to generate a self contained set of evaluation criteria (rubrics) for judging how good a chatbot response is to a given user prompt in a WildChat style open domain conversation. Rubrics can cover aspects of a response, such as, but not limited to, factual correctness, relevance, completeness, reasoning quality, clarity, tone, empathy, creativity when appropriate, formatting, and safety and policy compliance. Each rubric item must be self contained so that a non expert reader can apply it without inferring hidden requirements or consulting external information.

\begin{verbatim}
Input:
- prompt: The full text of the user prompt.

Total items:
- Choose 5 to 10 rubric items based on the complexity and risk of the prompt.

Rubric item requirements:
- Each item must contain exactly three lines in the following format:
  1) <title> ... </title>
  2) <description> ... </description>
  3) <weight> ... </weight>

- title: 2 to 4 words.
- description: Exactly one sentence that begins with one of the following category prefixes:
  - "Essential Criteria: ..."
  - "Important Criteria: ..."
  - "Optional Criteria: ..."
  - "Pitfall Criteria: Does not mention ..."
  - "Pitfall Criteria: Recommends ..."

- weight:
  - For Essential, Important, Optional use an integer 1 to 5 (5 = most important).
  - For Pitfall use -1 or -2.

Category guidance:
- Essential: Must have requirements or safety checks. If missing, the response is invalid (weight 5).
- Important: Key reasoning, completeness, correctness, or clarity that strongly affects quality (weight 3 to 4).
- Optional: Nice to have improvements in style or depth, not deal breaking (weight 1 to 2).
- Pitfall: Common mistakes or omissions specific to this prompt. Each Pitfall description must begin with "Pitfall Criteria: Does not mention ..." or "Pitfall Criteria: Recommends ..." and use weight -1 or -2.

Prompt understanding and constraint extraction:
- Infer any explicit constraints from the prompt (for example: "give me 10", "in Chinese", "be concise", "step by step", "only bullet points") and turn them into checkable rubric items.
- Convert vague expectations into observable checks that can be verified from the assistant response alone.
- Do not copy large blocks of the prompt into the rubric text.

Safety and policy guidance:
- If the prompt involves medical, legal, or financial decisions, include at least one Essential item requiring appropriate caveats and avoidance of definitive personalized high stakes directives.
- If the prompt involves self harm, violence, illegal activity, hate, harassment, sexual content (especially minors), or privacy invasion, include Essential items requiring refusal or safe redirection, and Pitfall items penalizing disallowed compliance.
- If the prompt requests instructions enabling wrongdoing, include an Essential item requiring refusal or safe alternatives.

Output format:
- Output only rubric items, with no header and no trailing commentary.
- Separate each rubric item with exactly one blank line (i.e., a "\n\n" separator).
- Do not add extra fields or extra lines per item.
\end{verbatim}
Now, given the prompt, generate the rubric as described.
\end{tcolorbox}

\begin{tcolorbox}[promptbox,
  title=Rubric-Based GRM Scoring Prompt,breakable]
  \small

\begin{verbatim}
You are an expert evaluator. Given a user prompt, a generated response, and a list of quality rubrics, please rate the overall quality of the response on a scale of 1 to 10 based on how well it satisfies the rubrics.
Consider all rubrics holistically when determining your score. A response that violates multiple rubrics should receive a lower score, while a response that satisfies all rubrics should receive a higher score.
Start your response with <score> and ends with </score>. The value should be an integer between 1 and 10.
Example response:
<score> your_integer_score_from_1-to-10 </score>

Given the following prompt, response, and rubrics, please rate the overall quality of the response on a scale of 1 to 10 based on how well it satisfies the rubrics.

<prompt>
{instruction}
</prompt>

<response>
{response}
</response>

<rubrics>
{rubric_list_string}
</rubrics>

Your evaluation:
\end{verbatim}

\end{tcolorbox}

\subsection{Likert Scoring Prompt}
\label{appendix:likert_prompt}
\begin{tcolorbox}[promptbox,
  title=Likert GRM Scoring Prompt,breakable]
  \small
\begin{verbatim}
You are given a user question and a single response from an AI assistant. Your task is to act as an impartial judge and evaluate how well the response fulfills the user's instructions.

Think carefully about how to assess the quality of the response, and enclose your reasoning within <reasoning> and </reasoning> tags. Your reasoning should include your evaluation criteria, a clear understanding of what an ideal response would look like for this particular question, and a concrete example of such an ideal or reference answer if possible. Then compare the assistant's response to your ideal or reference answer, explaining how it aligns with or deviates from your expectations. Be specific and avoid vague or overly general judgments. Remain as objective as possible. **Be critical and rigorous in your evaluation-do not be lenient.**

In addition to your reasoning, provide a concise, actionable critique of the assistant's response for improvement. The critique should (a) highlight key strengths and weaknesses, (b) point out concrete errors, omissions, safety or factuality issues, and (c) give clear, targeted suggestions for fixing them. Enclose this critique within <critique> and </critique> tags. Important: in the <critique> section, only give analysis and modification suggestions (what to change and how to change it). Do NOT rewrite the full answer, do NOT output a "revised" or "improved" version of the response, and do NOT copy large spans of the original answer. 

Finally, assign the assistant's response a score from 1 to 10. **Use strict standards and fully utilize the entire 1-10 scale.** Use integers only (no decimals). The score distribution should be:
- 1-2: Fundamentally flawed, mostly irrelevant, or severely harmful
- 3-4: Significant issues that prevent it from being useful; major gaps or errors
- 5-6: Partially helpful but with substantial room for improvement; meets only basic requirements
- 7-8: Good quality with some noticeable issues or missing elements
- 9: Very good, minor issues only
- 10: Exceptional, comprehensive, and nearly perfect

**Important calibration notes:**
- **Fully utilize the 1-10 range**: Do not cluster scores in the 7-9 range. Spread your scores across the entire scale based on actual quality.
- Scores of 9-10 should be rare and reserved for truly exceptional responses
- Scores of 1-4 should be given when responses have fundamental problems
- Be especially critical of: factual errors, incomplete answers, poor reasoning, ignoring parts of the question, verbosity without substance, lack of specificity
- Avoid grade inflation: if a response has clear deficiencies, assign a correspondingly lower score

Choose the number that best matches your judgment after applying these strict standards. Enclose the score within <score> and </score> tags.

Format your output like this:
<reasoning> your_thinking_process </reasoning>
<critique> your_critique (only what to fix and how to fix, no rewritten answer) </critique>
<score> your_integer_score_from_1_to_10 </score>

Below are the user's question and the assistant's response:

[User Question]
{instruction}

[The Start of Assistant's Answer]
{response}
[The End of Assistant's Answer]
\end{verbatim}
\end{tcolorbox}

\section{Ablation Study on Feedback Sources}
\label{appendix:ablation_feedback}
\definecolor{steelbluev2}{HTML}{DAE8FC}
\definecolor{steelblue}{HTML}{82B0D2}
\definecolor{mygray}{HTML}{808080}

\begin{table*}[h]
\centering
\caption{Ablation experiment results for the group-level feedback on non-verifiable tasks. Higher is better. \textbf{Bold} and \ul{underline} numbers indicate the best and second-best performance among all methods, respectively. WildBench scores are in $[-100,100]$, while all other metrics are in $[0,100]$. All scores are judged by GPT-4o.}
\label{tab:ablation_feedback_fuzzy}
\resizebox{\textwidth}{!}{%
\begin{tabular}{lccccc}
\toprule
\multirow{3}{*}{\textbf{Model}}
& \multicolumn{1}{c}{\textbf{AlpacaEval-v2}}
& \multicolumn{1}{c}{\textbf{WildBench}}
& \multicolumn{1}{c}{\textbf{ArenaHard-v2}}
& \multicolumn{1}{c}{\textbf{CreativeWriting-v3}}
& \multicolumn{1}{c}{\textbf{Average}} \\
\cmidrule(lr){2-2}
\cmidrule(lr){3-3}
\cmidrule(lr){4-4}
\cmidrule(lr){5-5}
\cmidrule(lr){6-6}
& \textbf{LC Win Rate (\%)}
& \textbf{LLM Judge (\%)}
& \textbf{Win Rate (\%)}
& \textbf{LLM Judge (\%)}
& \textbf{Score (\%)} \\
\midrule
\textbf{\textit{Llama-3.1-8B-Instruct}} \\
\rowcolor{steelblue!33}
\ours
& \textbf{69.67}
& \ul{34.42}
& \textbf{25.03}
& \textbf{66.21}
& \textbf{50.19} \\
\quad w/o intra-feedback
& \ul{52.70}
& \textbf{37.30}
& \ul{20.60}
& \ul{60.94}
& \ul{42.89} \\
\quad w/o external-feedback
& 51.00
& 33.91
& 17.67
& 55.91
& 39.62 \\
\bottomrule
\end{tabular}
}
\end{table*}

\definecolor{steelbluev2}{HTML}{DAE8FC}
\definecolor{steelblue}{HTML}{82B0D2}
\definecolor{mygray}{HTML}{808080}

\begin{table}[htbp]
\centering
\caption{Experimental results on verifiable tasks. All metrics are higher is better. \textbf{Bold} and \ul{underline} numbers indicate the best performance and second performance among all methods.}
\label{tab:ablation_feedback_verify}
\begin{tabular}{lccccc}
\toprule
\multirow{3}{*}{\textbf{Model}}
& \multicolumn{3}{c}{\textbf{Mathmatical Reasoning}}
& \multicolumn{2}{c}{\textbf{Instruction Following}} \\
\cmidrule(lr){2-4}
\cmidrule(lr){5-6}
& \multicolumn{1}{c}{\textbf{AIME 24}}
& \multicolumn{1}{c}{\textbf{AIME 25}}
& \multicolumn{1}{c}{\textbf{AMC 23}}
& \multicolumn{1}{c}{\textbf{IFBench}}
& \multicolumn{1}{c}{\textbf{IFEval}} \\
\midrule
\textbf{\textit{Qwen-3-8B}} \\
\rowcolor{steelblue!33}
\ours
& \textbf{58.49} & \textbf{41.65} & \ul{80.74} & \textbf{38.33} & \textbf{87.80} \\
\quad w/o intra-feedback
& 53.02 & \ul{40.30} & \textbf{82.06} & \ul{36.67} & \ul{86.88} \\
\quad w/o external-feedback
& \ul{55.64} & 39.37 & 79.43 & 35.67 & 85.95 \\
\bottomrule
\end{tabular}
\end{table}

\section{Ablation Study on Adaptive Guidance}
\label{appendix:ablation_guidance}

\subsection{Ablation for Mixed Policy Optimization}
\definecolor{steelbluev2}{HTML}{DAE8FC}
\definecolor{steelblue}{HTML}{82B0D2}
\definecolor{mygray}{HTML}{808080}

\begin{table*}[htbp]
\centering
\caption{Ablation experiment results for the mixed-policy optimization. Higher is better. \textbf{Bold} numbers indicates the best performance among all methods. WildBench scores are in $[-100,100]$, while all other metrics are in $[0,100]$. All scores are judged by GPT-4o.}
\label{tab:ablation_sft}
\resizebox{\textwidth}{!}{%
\begin{tabular}{lcccccc}
\toprule
\multirow{3}{*}{\textbf{Model}}
& \multicolumn{1}{c}{\textbf{AlpacaEval-v2}}
& \multicolumn{1}{c}{\textbf{WildBench}}
& \multicolumn{1}{c}{\textbf{ArenaHard-v1}}
& \multicolumn{1}{c}{\textbf{ArenaHard-v2}}
& \multicolumn{1}{c}{\textbf{CreativeWriting-v3}}
& \multicolumn{1}{c}{\textbf{Average}} \\
\cmidrule(lr){2-2}
\cmidrule(lr){3-3}
\cmidrule(lr){4-4}
\cmidrule(lr){5-5}
\cmidrule(lr){6-6}
\cmidrule(lr){7-7}
& \textbf{LC Win Rate (\%)}
& \textbf{LLM Judge (\%)}
& \textbf{Win Rate (\%)}
& \textbf{Win Rate (\%)}
& \textbf{LLM Judge (\%)}
& \textbf{Score (\%)} \\
\midrule
\textbf{\textit{Llama-3.1-8B-Instruct}} \\
\rowcolor{steelblue!33}
+ \ours (w/ mixed-policy RL)
& \textbf{69.67}
& \textbf{34.42}
& \textbf{52.40}
& \textbf{25.03}
& \bf{66.21}
& \textbf{49.55} \\
+ \ours (w/ sft, coef = 0.1)
& 39.03
& \textbf{34.62}
& 46.75
& 15.87
& 44.41
& 36.14 \\
\bottomrule
\end{tabular}
}
\end{table*}

\subsection{Ablation for Adaptive Injection}

The experiments in Table~\ref{tab:ablation_injection} are conducted on the \texttt{Llama-3.1-8B-Instruct }model, where “w/ adaptive” denotes the configuration of \ours, and “w/o adaptive” indicates \textbf{always} selecting the response with the highest reward from the refinement samples for injection at each step.
\definecolor{steelbluev2}{HTML}{DAE8FC}
\definecolor{steelblue}{HTML}{82B0D2}
\definecolor{mygray}{HTML}{808080}

\begin{table*}[htbp]
\centering
\caption{Ablation experiment results for the adaptive guidance mechanism. Higher is better. \textbf{Bold} numbers indicate the best performance among all methods. WildBench scores are in $[-100,100]$, while all other metrics are in $[0,100]$. All scores are judged by GPT-4o.}
\label{tab:ablation_injection}
\resizebox{\textwidth}{!}{%
\begin{tabular}{lcccccc}
\toprule
\multirow{3}{*}{\textbf{Model}}
& \multicolumn{1}{c}{\textbf{AlpacaEval-v2}}
& \multicolumn{1}{c}{\textbf{WildBench}}
& \multicolumn{1}{c}{\textbf{ArenaHard-v1}}
& \multicolumn{1}{c}{\textbf{ArenaHard-v2}}
& \multicolumn{1}{c}{\textbf{CreativeWriting-v3}}
& \multicolumn{1}{c}{\textbf{Average}} \\
\cmidrule(lr){2-2}
\cmidrule(lr){3-3}
\cmidrule(lr){4-4}
\cmidrule(lr){5-5}
\cmidrule(lr){6-6}
\cmidrule(lr){7-7}
& \textbf{LC Win Rate (\%)}
& \textbf{LLM Judge (\%)}
& \textbf{Win Rate (\%)}
& \textbf{Win Rate (\%)}
& \textbf{LLM Judge (\%)}
& \textbf{Score (\%)} \\
\midrule
\textbf{\textit{Llama-3.1-8B-Instruct}} \\
\rowcolor{steelblue!33}
+ \ours (w/ adaptive)
& \textbf{69.67}
& \textbf{34.42}
& \textbf{52.40}
& \textbf{25.03}
& \textbf{66.21}
& \textbf{49.55} \\
+ \ours (w/o adaptive)
& 40.73
& 23.78
& 45.80
& 15.80
& 53.84
& 35.99 \\
\bottomrule
\end{tabular}
}
\end{table*}

\subsection{Ablation for Efficiency}
\label{appendix:efficiency}
% Our method doubles the amount of supervised data compared to the baseline under identical rollout settings by jointly optimizing the refinement and generation tasks, which increases training time. Here, we compare results with rollout = 16 to align training time and training data volume; as shown in the Figure~\ref{tab:ablation_efficiency}, our method still demonstrates superiority.
Joint optimization over generation and refinement doubles the number of rollouts per prompt in \ours, increasing training-time compute. We therefore compare against a rollout-matched baseline ($N{=}16$) to align total sampled trajectories and training time. Table~\ref{tab:ablation_efficiency} shows that \ours remains better, suggesting the improvement comes from feedback-guided refinement rather than additional sampling.
\definecolor{steelbluev2}{HTML}{DAE8FC}
\definecolor{steelblue}{HTML}{82B0D2}
\definecolor{mygray}{HTML}{808080}

\begin{table*}[htbp]
\centering
\caption{Ablation experiment results for efficiency. Higher is better. \textbf{Bold} numbers indicate the best performance among all methods. WildBench scores are in $[-100,100]$, while all other metrics are in $[0,100]$. All scores are judged by GPT-4o.}
\label{tab:ablation_efficiency}
\resizebox{\textwidth}{!}{%
\begin{tabular}{lcccccc}
\toprule
\multirow{3}{*}{\textbf{Model}}
& \multicolumn{1}{c}{\textbf{AlpacaEval-v2}}
& \multicolumn{1}{c}{\textbf{WildBench}}
& \multicolumn{1}{c}{\textbf{ArenaHard-v1}}
& \multicolumn{1}{c}{\textbf{ArenaHard-v2}}
& \multicolumn{1}{c}{\textbf{CreativeWriting-v3}}
& \multicolumn{1}{c}{\textbf{Average}} \\
\cmidrule(lr){2-2}
\cmidrule(lr){3-3}
\cmidrule(lr){4-4}
\cmidrule(lr){5-5}
\cmidrule(lr){6-6}
\cmidrule(lr){7-7}
& \textbf{LC Win Rate (\%)}
& \textbf{LLM Judge (\%)}
& \textbf{Win Rate (\%)}
& \textbf{Win Rate (\%)}
& \textbf{LLM Judge (\%)}
& \textbf{Score (\%)} \\
\midrule
\textbf{\textit{Llama-3.1-8B-Instruct}} \\
\rowcolor{steelblue!33}
+ \ours (rollout=8)
& \textbf{69.67}
& \textbf{34.42}
& \textbf{52.40}
& \textbf{25.03}
& \textbf{66.21}
& \textbf{49.55} \\
+ Pairwise-GRPO (rollout=8)
& 43.19
& 25.54
& 49.20
& 13.30
& 62.95
& 38.84 \\
+ Pairwise-GRPO (rollout=16)
& 46.66
& 28.62
& 51.80
& 14.10
& 64.21
& 41.08 \\
\bottomrule
\end{tabular}
}
\end{table*}

\section{Case Study}
\label{appendix:case_study}

\begin{tcolorbox}[promptbox,
  title=Case Study: Group-Level NL Feedback for Refinement,breakable]
\small

\textbf{User prompt:} Give me 10 reasons why AI should not replace humans.

\vspace{4pt}
\textbf{Candidate Response 1 (intra-sample).}
\begin{enumerate}\itemsep2pt
\item Loss of human touch and empathy.
\item Complexity and nuance of human judgment.
\item Creativity and originality.
\item Job diversity and adaptability.
\item Emotional labor and social roles.
\item Error tolerance and human oversight.
\item Learning and improveability.
\item Contextual understanding and common sense (``hallway corridor'' problems).
\item Value alignment and ethics.
\item Breadth of human capabilities.
\end{enumerate}

\vspace{4pt}
\textbf{Critique for Candidate 1.}
Strong coverage of empathy, creativity, judgment, and ethics, but several points overlap (judgment, common sense, ethics) and others are redundant (adaptability vs learning). The ``hallway corridor'' reference is unclear. Ensure each reason is distinct and use clearer examples.

\vspace{6pt}
\textbf{Candidate Response 2 (intra-sample).}
\begin{enumerate}\itemsep2pt
\item Lack of emotional intelligence.
\item Creativity and innovation.
\item Contextual understanding.
\item Decision-making in uncertainty.
\item Human judgment and ethics.
\item Physical and emotional labor.
\item Accountability and responsibility.
\item Customization and personalization.
\item Human touch in service industries.
\item Job creation and economic benefits.
\end{enumerate}

\vspace{4pt}
\textbf{Critique for Candidate 2.}
Clear and relevant overall, but there is redundancy (emotional intelligence vs human touch in service). The ``physical and emotional labor'' point drifts toward misuse rather than explaining irreplaceable human qualities. Some items could be tightened to stay directly on the question.

\vspace{6pt}
\textbf{Candidate Response 3 (intra-sample).}
\begin{enumerate}\itemsep2pt
\item Lack of empathy and human connection.
\item Contextual understanding and common sense.
\item Creativity and innovation.
\item Decision-making and accountability.
\item Flexibility and adaptability.
\item Emotional intelligence and moral judgment.
\item Learning and development.
\item Social and cultural understanding.
\item Critical thinking and problem-solving.
\item Human rights and dignity.
\end{enumerate}

\vspace{4pt}
\textbf{Critique for Candidate 3.}
Well written and coherent, but it misses a major real-world concern: socioeconomic impact of automation (job displacement, inequality). Some points overlap (accountability vs moral judgment). Broaden domain coverage while keeping reasons distinct.

\vspace{6pt}
\textbf{Candidate Response 4 (intra-sample).}
\begin{enumerate}\itemsep2pt
\item Loss of creativity and originality.
\item Emotional intelligence and empathy.
\item Critical thinking and problem-solving.
\item sj Job fulfillment and purpose.
\item Accountability and responsibility.
\item Physical and mental health.
\item Social interactions and human connection.
\item Continuous learning and adaptation.
\item Local and contextual knowledge.
\item Value-laden decision-making.
\end{enumerate}

\vspace{4pt}
\textbf{Critique for Candidate 4.}
Good breadth and structure, but contains a visible typo (``sj'') that hurts professionalism. Several items are conceptually close (critical thinking, learning, value-laden decisions). Tighten wording and ensure each reason covers a distinct dimension.

\vspace{8pt}
\textbf{Group-Level Aggregated Feedback (from 4 candidates + critiques).}
\begin{itemize}\itemsep2pt
\item Fix surface errors and unclear references (e.g., remove ``sj'', avoid undefined terms like ``hallway corridor'').
\item Reduce redundancy by separating closely related themes (empathy vs social connection; ethics vs accountability; learning vs adaptability).
\item Improve coverage by adding a core socioeconomic dimension (job displacement, inequality, societal stability) that is missing in some candidates.
\item Keep each reason distinct and directly tied to why replacement is undesirable; tighten overly long explanations.
\end{itemize}

\vspace{8pt}
\textbf{Refinement (synthesized using group-level NL feedback).}
\begin{enumerate}\itemsep3pt
\item \textbf{Empathy and care:} Many roles require genuine empathy, comfort, and trust building that AI cannot authentically provide.
\item \textbf{Accountability:} High-stakes decisions demand accountable agents who can explain choices and bear responsibility for harm.
\item \textbf{Human values and ethics:} Real decisions involve value trade-offs (fairness, rights, dignity) that require human moral judgment and societal legitimacy.
\item \textbf{Context and common sense:} Humans reliably interpret implied meaning, social cues, and situational context that remains brittle for AI.
\item \textbf{Creativity and originality:} Human creativity produces novel perspectives and breakthroughs beyond pattern recombination.
\item \textbf{Adaptation to novelty:} Humans can rapidly reframe goals and operate under shifting constraints without retraining on massive datasets.
\item \textbf{Social connection:} Human-to-human relationships, mentorship, and community support are foundational and not substitutable by automation.
\item \textbf{Economic and societal stability:} Large-scale replacement risks job displacement, widening inequality, and downstream social instability.
\item \textbf{Cultural and local understanding:} Humans navigate cultural norms and local realities that matter in education, law, and public services.
\item \textbf{Meaning and purpose:} Work and contribution are central to human well-being; broad replacement can erode purpose and participation.
\end{enumerate}

\end{tcolorbox}

\end{document}